\documentclass[10pt,twocolumn,letterpaper]{article}
\usepackage[pagenumbers]{cvpr}

\definecolor{db}{RGB}{144,218,130}
\definecolor{b}{RGB}{229,254,82}
\definecolor{lb}{RGB}{255,253,212}
\definecolor{antiquewhite}{rgb}{0.98, 0.92, 0.84}

\usepackage{soul}

\usepackage{placeins}
\usepackage{graphicx}
\usepackage{multirow}
\usepackage{amsfonts}
\usepackage{amsthm}
\usepackage{mathrsfs}
\usepackage[title]{appendix}
\usepackage{textcomp}
\usepackage{manyfoot}
\usepackage{algorithm}
\usepackage{algorithmicx}
\usepackage{algpseudocode}
\usepackage{listings}

\usepackage{tabularx}
\usepackage{multicol}
\usepackage{colortbl}
\usepackage{array}

\usepackage{makecell}
\usepackage{stfloats}

\definecolor{cvprblue}{rgb}{0.21,0.49,0.74}
\usepackage[pagebackref,breaklinks,colorlinks,allcolors=cvprblue]{hyperref}

\def\paperID{} 
\def\confName{CVPR}
\def\confYear{2026}

\title{PhysDepth: Plug-and-Play Physical Refinement for Monocular Depth Estimation in Challenging Environments}

\author{
Kebin Peng\textsuperscript{1} \quad Haotang Li\textsuperscript{2} \quad Zhenyu Qi\textsuperscript{2} \quad Huashan Chen\textsuperscript{3,4} \\ Zi Wang\textsuperscript{5} \quad Wei Zhang\textsuperscript{5} \quad Sen He\textsuperscript{2} \quad Huanrui Yang\textsuperscript{2} \quad Qing Guo\textsuperscript{6} \vspace{0.3em} \\
{\normalsize \textsuperscript{1} Department of Computer Science, East Carolina University, Greenville, NC, USA} \\
{\normalsize \textsuperscript{2} Department of Electrical and Computer Engineering, The University of Arizona, Tucson, AZ, USA} \\
{\normalsize \textsuperscript{3} Institute of Information Engineering, Chinese Academy of Sciences, Beijing, China} \\
{\normalsize \textsuperscript{4} School of Cyber Security, University of Chinese Academy of Sciences, Beijing, China} \\
{\normalsize \textsuperscript{5} School of Computer and Cyber Sciences, Augusta University, Augusta, GA, USA} \\
{\normalsize \textsuperscript{6} VCIP, CS, Nankai University, China}
}

\begin{document}
\maketitle
\begin{abstract}
State-of-the-art monocular depth estimation (MDE) models often struggle in challenging environments, primarily because they overlook robust physical information.
To demonstrate this, we first conduct an empirical study by computing the covariance between a model's prediction error and atmospheric attenuation.
We find that the error of existing SOTAs increases with atmospheric attenuation.
Based on this finding, we propose PhysDepth, a plug-and-play framework that solves this fragility by infusing physical priors into modern SOTA backbones.
PhysDepth incorporates two key components: a Physical Prior Module (PPM) that leverages Rayleigh Scattering theory to extract robust features from the high-SNR red channel, and a physics-derived Red Channel Attenuation Loss (RCA) that enforces model to learn the Beer-Lambert law.
Extensive evaluations demonstrate that PhysDepth achieves SOTA accuracy in challenging conditions. 
\end{abstract}    
\section{Introduction}
\label{sec:intro}

Monocular depth estimation (MDE) is a fundamental task for autonomous driving \cite{park2021pseudo,wang2019pseudo} and robotics \cite{zhou2019does,dong2022towards}.
While recent self-supervised methods have advanced significantly, their performance remains inherently limited by a primarily data-driven paradigm \cite{gasperini2023robust,rajapaksha2024deep,liu2021self,vankadari2020unsupervised,wang2021regularizing,vankadari2023sun,zheng2023steps,godard2019digging,vankadari2024dusk,spencer2020defeat}.
This reliance forces models to learn brittle, non-physical heuristics (e.g., 'finer texture: closer' or 'brighter lights: closer') rather than robust physical principles.
These spurious correlations often struggle in challenging environments, resulting in suboptimal results.

The fundamental problem is that existing SOTA methods \textit{overlook robust physical information}.

\begin{figure}[t]
    \includegraphics[width=\linewidth]{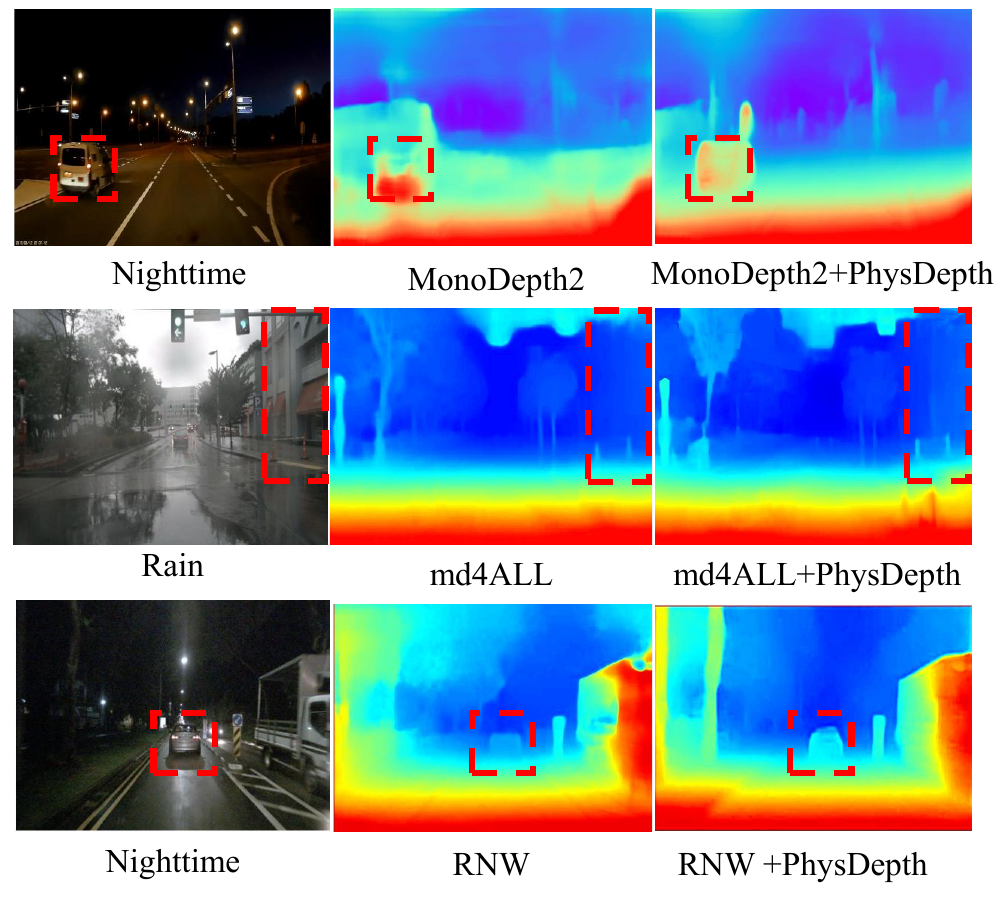}
    \vspace{-8mm}
    \captionof{figure}{\textbf{Plug-and-Play of the PhysDepth.} we show that adding PhysDepth to a baselines like MonoDepth2 \cite{godard2019digging}, md4ALL\cite{gasperini2023robust}, and RNW \cite{wang2021regularizing} effectively corrects common failure modes.
    }
    \label{fig:example-vis-results}
    \vspace{-6mm}
\end{figure}

Motivated by this, we develop PhysDepth, a novel self-supervised MDE framework that infuses physical priors, Rayleigh Scattering theory and Beer-Lambert law, into modern SOTA backbones (e.g., ViT and ConvNeXt).
Specifically, PhysDepth enhances base MDE model with two key components: a \textit{Physical Prior Module (PPM)} and a \textit{Red Channel Attenuation Loss (RCA)}. 
The PPM, guided by Rayleigh Scattering, extracts robust features from the high-SNR red channel. 
The RCA loss, derived from the Beer-Lambert law, enforces a physically-consistent relationship between this attenuation signal and the final depth. 

As illustrated in Figure \ref{fig:example-vis-results}, SOTA MDE models, including MonoDepth2 \cite{godard2019digging}, md4ALL\cite{gasperini2023robust}, and RNW \cite{wang2021regularizing} fail in challenging environments, producing hazardous artifacts from real-world conditions like road reflections and glare.

By plugging in PhysDepth, enhanced MDE models can produce visually clean and accurate results.   
Furthermore, our quantitative analysis in Table \ref{tab:plug-in} shows that when we plug
PhysDepth into existing SOTA backbones, their performance has been significantly improved.

In summary, the key contributions of this paper include:

\begin{itemize}
    \item We find that existing SOTA MDE models overlook physical information. To prove this, we conduct an empirical study that measures the correlation between atmospheric attenuation and the model's error metric.
    \item We propose PhysDepth, a novel plug-and-play framework that addresses this fragility by explicitly integrating physical priors (Rayleigh Scattering and Beer-Lambert law) into diverse modern backbones.
    \item We introduce a Physical Prior Module (PPM) and a physics-derived Red Channel Attenuation Loss (RCA), which together form a portable plugin for different MDE architectures.
    \item Extensive evaluations demonstrate that PhysDepth achieves SOTA performance in challenging conditions, and prove the generality of our plug-and-play approach.
\end{itemize}

\section{Related Work}
\label{sec:related work}

Depth estimation under challenging conditions (e.g., rain or night) has attracted increased attention recently.
Some SOTA methods focus on feature representation and regularizers. 
For example, 
Zhang et al. \cite{zhang2025lightweight} leverages CoMoGAN to transform daytime images into challenging scenes, thereby circumventing the need for a separate nighttime dataset.
Yan et al. \cite{Yan_2025_CVPR} also use synthetic adverse data to train their model.
Zhang et al \cite{Zhang_2025_ICCV} use CLIP and DINO under contrastive language guidance for dark images.
Chung et al. \cite{Chung_2025_ICCV} proposed a test-time adaptation for challenging condition depth estimation.
DeFeat-Net \cite{spencer2020defeat} uses a cross-domain dense feature representation.
RNW \cite{wang2021regularizing} uses image enhancement and a GAN-based regularizer. 
Cong et al. \cite{10696933} introduce a structure-regularized framework cropped multi-scale consistency loss for low visibility and illumination conditions.
Meanwhile, other SOTA methods focus on domain adaptation networks.
ADIDS \cite{liu2021self} uses two networks for day-time and nighttime images. 
Vankadari et al. \cite{vankadari2020unsupervised} and ITDFA \cite{zhao2022unsupervised}
both adapted a domain adaptation network trained on daytime images to function in dark environments. 
Gasperini et al. \cite{gasperini2023robust} also convert challenging environment images to daytime conditions using a translation model.
WSGD \cite{vankadari2023sun} adds denoising into domain adaptation network.
Hou et al. \cite{10539960} introduce an illumination compensation PoseNet for lighting adaptation, and a cross-layer adaptive fusion module to enhance feature complementarity.
Sharma et al. \cite{sharma2020nighttime} developed a translation network to create nighttime stereo images from daytime versions, then trained on a stereo network.
Also, some SOTA methods use additional information.
Lu et al. \cite{lu2021alternative} combined thermal and RGB data to enhance the accuracy of nighttime depth estimation. 
R4dyn \cite{gasperini2021r4dyn} uses radar at nighttime.

\section{PhysDepth: Motivation \& Method}
\label{sec:framework}

\subsection{Empirical Study, Motivation, and Base Model}

\begin{figure}[ht]
    \centering
    \includegraphics[width=\linewidth]{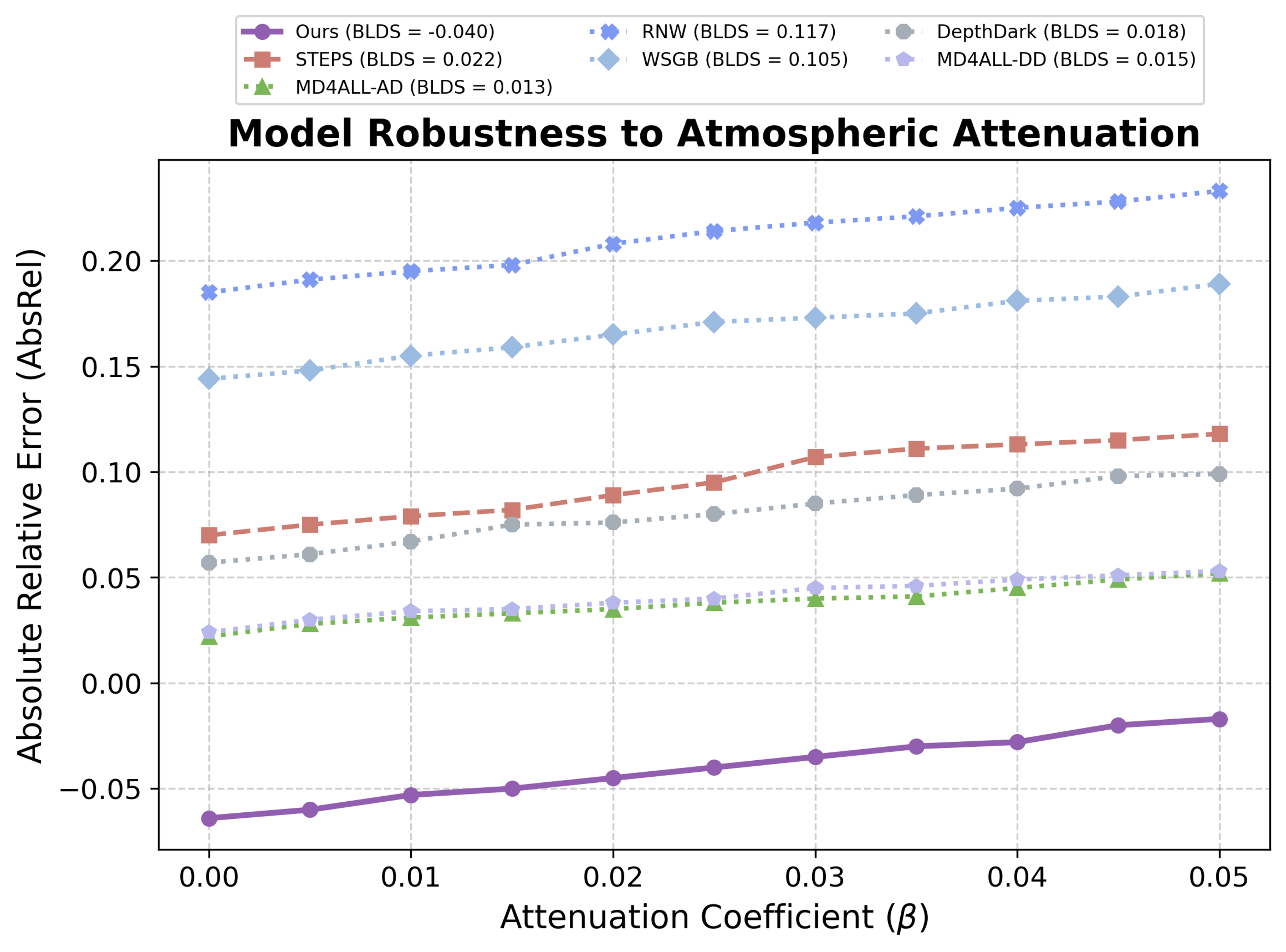}
    \vspace{-6mm}
    \caption{\textbf{Quantitative Analysis of Robustness to Atmospheric Attenuation.} Quantitative comparison of model robustness against increasing atmospheric attenuation ($\beta$). Our method (purple line) achieves the lowest Absolute Relative Error (AbsRel) and the flattest performance curve.}
    \label{fig:BLDS}
    \vspace{-5mm}
\end{figure}

\begin{figure*}[t]
	\centering
	\includegraphics[width=\linewidth]{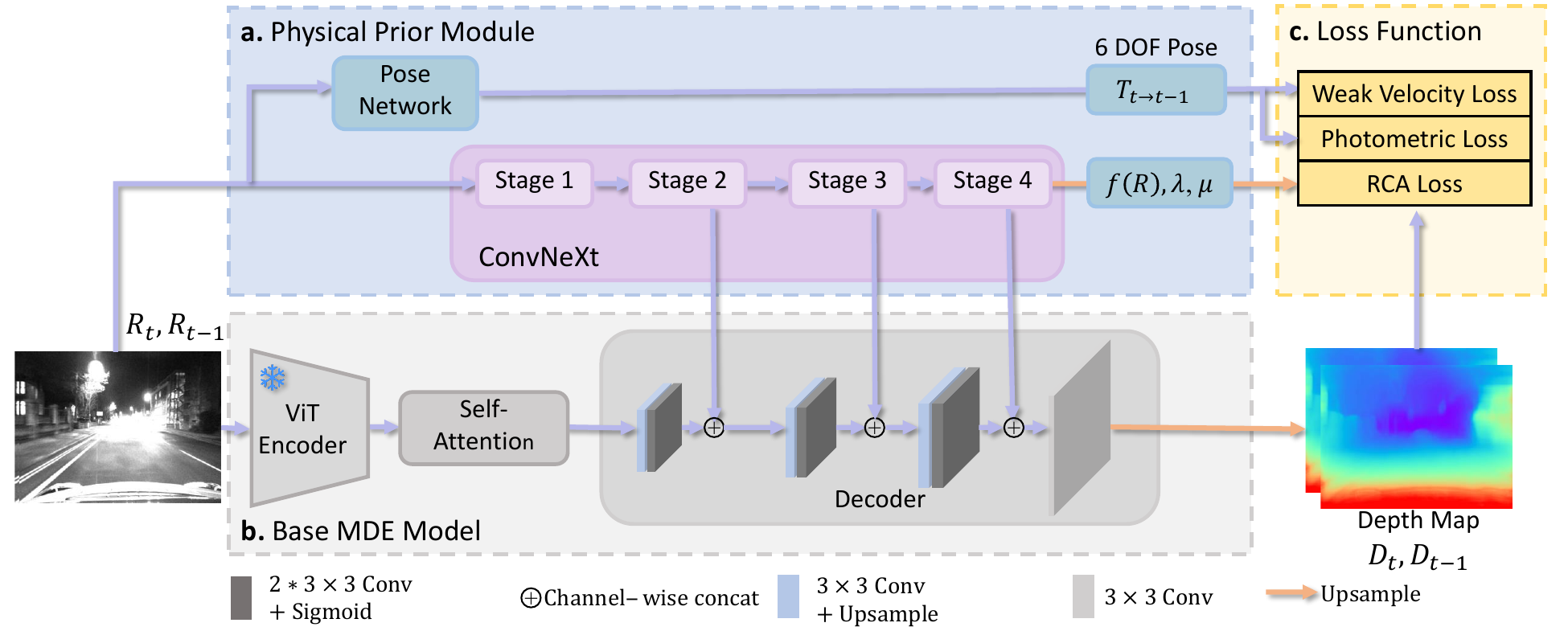}
	\caption{\textbf{PhysDepth Architecture} contains three parts:\textbf{(a)} The Physical Prior Module (PPM), which uses a ConvNeXt backbone to extract hierarchical features from the red channels (Rt, Rt-1). \textbf{(b)} The Base MDE Model, which uses a ViT Encoder and a Decoder to predict depth from the full image. The PPM's features are fused into the Base Decoder at multiple scales. \textbf{(c)} The Loss Function, which includes our RCA Loss that supervises the PPM. }
    \label{fig:network}
    \vspace{-5mm}
\end{figure*}

We argue that existing SOTA methods overlook robust physical information. 
To prove this, we conduct an empirical study that measures the covariance $Cov$ between a model's error metric and a controlled, physically-plausible injection of atmosphere attenuation $\beta$ \cite{he2010single}.

This study uses a test set containing challenging condition images $I$ and corresponding ground-truth depth maps $D_{gt}$. We generate a sequence of perturbed images by applying the volumetric scattering model \cite{he2010single,tan2008visibility,10.1145/1399504.1360671,narasimhan2000chromatic,narasimhan2002vision} in Equation \ref{eq:blds} :

\begin{equation}
\label{eq:blds}
    I'(x) = J(x) \cdot t(x) + A \cdot (1 - t(x))
\end{equation}

We use algorithm \ref{alg:blds} in supplemental material to compute $Cov$, and we define the parameters as follows:
\begin{itemize}
    \item Scene Radiance $J(x)$: We approximate the clear-night radiance using the input image: $J(x) \approx I(x)$.
    \item Airlight $A$: This term represents ambient light scattered into the camera path. Critically, for a nocturnal scene, we fix a neutral, dark airlight, representative of ambient light pollution or environmental glow, such as $A = [0.1, 0.1, 0.1]$ in RGB space. This correctly simulates the "black-point" elevation observed in low-visibility conditions (e.g., night).
    \item Transmission $t(x)$: The transmission is governed by the Beer-Lambert law, $t(x) = \exp(-\beta \cdot D_{gt}(x))$, where $\beta$ is the extinction coefficient that controls the density of the attenuation (e.g., light fog or mist).
    \item Perturbation Vector $\mathbf{B}$: We define a vector of $N$ extinction coefficients, $\mathbf{B} = [\beta_0, \beta_1, ..., \beta_N]$. A typical range is $\beta_i \in [0.0, 0.05]$, where $\beta_0 = 0.0$ represents the original, clear-night image.
    \item Error function $\text{Error}$: We use  $Abs Rel$ here, a "lower-is-better" metric. 
\end{itemize}

\textbf{Interpretation of the covariance.} As shown in Figure \ref{fig:BLDS}, 
Our PhysDepth (purple solid line) achieves the lowest AbsRel error across the entire range of attenuation coefficients, consistently and significantly outperforming all competing models.
As $\beta$ increases, the error of all models rises.
However, PhysDepth exhibits the flattest performance curve, indicating that its performance degrades the least as conditions worsen. 
This superior stability is also quantified by the $Cov$ score shown in the legend; our method achieves the best (lowest) score of -0.040, while all other methods have positive (worse) scores, confirming its exceptional robustness to challenging environments.
Also, PhysDepth's $Cov < 0$ means that it has successfully learned the physical priors, treating attenuation as a valid, independent signal for depth.
SOTAs such as STEPS \cite{zheng2023steps}, DepthDark \cite{zeng2025depthdark}, md4ALL-AD \cite{gasperini2023robust}, md4ALL-DD \cite{gasperini2023robust}, their $Cov \approx 0.0$ (No Correlation) means each model's performance is decoupled from attenuation, indicating it \textit{only rely on other cues}, such as semantic context or geometry.
For other methods like WSGB \cite{vankadari2023sun} and RNW \cite{wang2021regularizing}, their $Cov > 0$ (Positive Correlation) indicates that each model treats attenuation as noise that occludes its primary cues.
This empirical study motivates us to design PhysDepth based on robust physical information.

Figure \ref{fig:network} illustrates the design of PhysDepth.
We select a modern Transformer-based architecture as the encoder in our primary Base MDE Models (Figure \ref{fig:network} - b).
We also follow Piccinelli et al. \cite{piccinelli2024unidepth} to employ both ConvNeXt-T \cite{liu2022convnet}  and MiDaS \cite{ranftl2020towards} to serve as the encoder in the base model. 

\subsection{Physical Prior Module}
\label{sec:PPM}

Rayleigh Scattering theory depicts the relationship between wavelength and the amount of scattered light (Equation \ref{eq:Rayleigh scattering}, where $ I_\text{s} $ is the intensity of scattered light, and $ \lambda $ is the wavelength). Specifically, red light, due to its longer wavelength, is less likely to be scattered than blue and green light. Thus, \emph{red channel image is more likely to preserve details for MDE, especially under challenging lighting conditions.}

\vspace{-3mm}
\begin{equation}
    \label{eq:Rayleigh scattering}
    I_\text{s} \propto \frac{1}{\lambda^4}
\end{equation}

We start by conducting experiments to verify the relationships between image channels and the accuracy of MDE results.
Particularly, RGB images from RobotCar-Night dataset \cite{RobotCarDatasetIJRR} are replaced by their single-channel images and two-channel images for comparison, respectively.
We use the base MDE model, Pose Network, and Photometric Loss in Figure \ref{fig:network} as a baseline.
Table \ref{tab:abl_channel_RobotCar_Night_baseline} shows that single-channel image settings outperform their two-channel counterparts, surpassing RGB image settings.
Notably, the red channel setting (Baseline+R) yields the best results. 

\begin{table}[ht]
    \centering
    \footnotesize
    \caption{\textbf{Using Different Channel Images from RobotCar-Night Dataset \cite{RobotCarDatasetIJRR}:} R means Red Channel, so do B (Blue Channel) and G (Green Channel), RGB. The best results are in \textbf{bold}. Base MDE model uses ViT-B/.}    \vspace{-2mm}
    \begin{tabular}{l|cccc}
    \toprule
    \multicolumn{1}{c|}{Methods} & \multicolumn{1}{c|}{Abs Rel} & \multicolumn{1}{c|}{Sq Rel} & \multicolumn{1}{c|}{RMSE} & \multicolumn{1}{c}{RMSE log} \\
    \hline
    \textbf{} & \multicolumn{4}{c}{lower's better} \\
    \hline
    
    Baseline+RGB & \multicolumn{1}{c|}{0.690} & \multicolumn{1}{c|}{8.215} & \multicolumn{1}{c|}{7.105} & \multicolumn{1}{c}{0.473} \\ \hline
    
    Baseline+B+R & \multicolumn{1}{c|}{0.673} & \multicolumn{1}{c|}{8.211} & \multicolumn{1}{c|}{7.105} & \multicolumn{1}{c}{0.473} \\ \hline
    
    Baseline+G+R & \multicolumn{1}{c|}{0.675} & \multicolumn{1}{c|}{8.212} & \multicolumn{1}{c|}{7.103} & \multicolumn{1}{c}{0.471} \\ \hline
    
    Baseline+B+G & \multicolumn{1}{c|}{0.671} & \multicolumn{1}{c|}{8.210} & \multicolumn{1}{c|}{7.104} & \multicolumn{1}{c}{0.469} \\ \hline
    
    Baseline+B & \multicolumn{1}{c|}{0.668} & \multicolumn{1}{c|}{8.209} & \multicolumn{1}{c|}{7.098} & \multicolumn{1}{c}{0.465} \\ \hline
    
    Baseline+G & \multicolumn{1}{c|}{0.667} & \multicolumn{1}{c|}{8.209} & \multicolumn{1}{c|}{7.098} & \multicolumn{1}{c}{0.463} \\ \hline
    
    \textbf{Baseline+R} & \multicolumn{1}{c|}{\cellcolor{db}\textbf{0.665}} & \multicolumn{1}{c|}{\cellcolor{db}\textbf{8.206}} & \multicolumn{1}{c|}{\cellcolor{db}\textbf{7.097}} & \multicolumn{1}{c}{\cellcolor{db}\textbf{0.461}} \\
    
    \bottomrule
    
    \end{tabular}
    \label{tab:abl_channel_RobotCar_Night_baseline}
\end{table}
 
The red channel's superior performance attributes to a two-fold physical advantage: \textbf{1)} Rayleigh Scattering: light is scattered by the atmospheric medium itself, even in clear air. This law states that the shorter-wavelength blue and green channels are scattered more than the red channel, degrading their signal integrity over distance. 
\textbf{2)} Signal-to-Noise Ratio (SNR): This fundamental scattering advantage is compounded by the realities of challenging-condition illumination. 
Artificial lighting (e.g., brake lights, streetlights) is mainly red/yellow.
As a result, the sensor's red channel receives a stronger and cleaner signal, while the blue and green channels are more likely to capture sensor noise. 
Furthermore, combining multiple channels can introduce additional noise \cite{hong2024you}, leading to inaccurate MDE results. 
Thus, we use \textbf{red channel images} for PhysDepth. 

To take full advantage of red channel images, we propose PPM (Figure \ref{fig:network} - a), which aims to enable backbone (e.g., ViT-B/32) and pose estimation network to \textit{explicitly} and \textit{comprehensively} learn the relationships between red channel values and depth values.

To implement PPM, we employ a modern hierarchical backbone, ConvNeXt \cite{liu2022convnet}, to extract image features and establish the correlation between red channel values and depth values (Figure \ref{fig:network} - a).
To enhance the robustness of the Base MDE Model, we inject the physical features from the PPM at multiple scales.
The intermediate feature maps from the ConvNeXt encoder stages (Stage 2, Stage 3, Stage 4) are fused with the feature maps of the Base MDE Decoder at corresponding resolutions via channel-wise concatenation.

The output of PPM is the predicted depth map from red channel image. It also serves as part of the inputs of RCA loss (See Section \ref{sec:rca loss}). 
Simultaneously, the predicted depth map from the decoder serves as the other part of RCA loss input.
To match the dimensions of the predicted depth map, the last layer of PPM's output requires an up-sampling layer.
One noteworthy difference between our PPM and a standard ConvNeXt encoder is that, while they both output the same dimension for depth estimation, PPM, in addition, outputs additional two channels for estimating $ \mu $ and $ \lambda $ for RCA loss in Equation \ref{eq:rca loss r and c}.

\begin{figure*}[ht]
	\centering
	\includegraphics[width=0.7\linewidth]{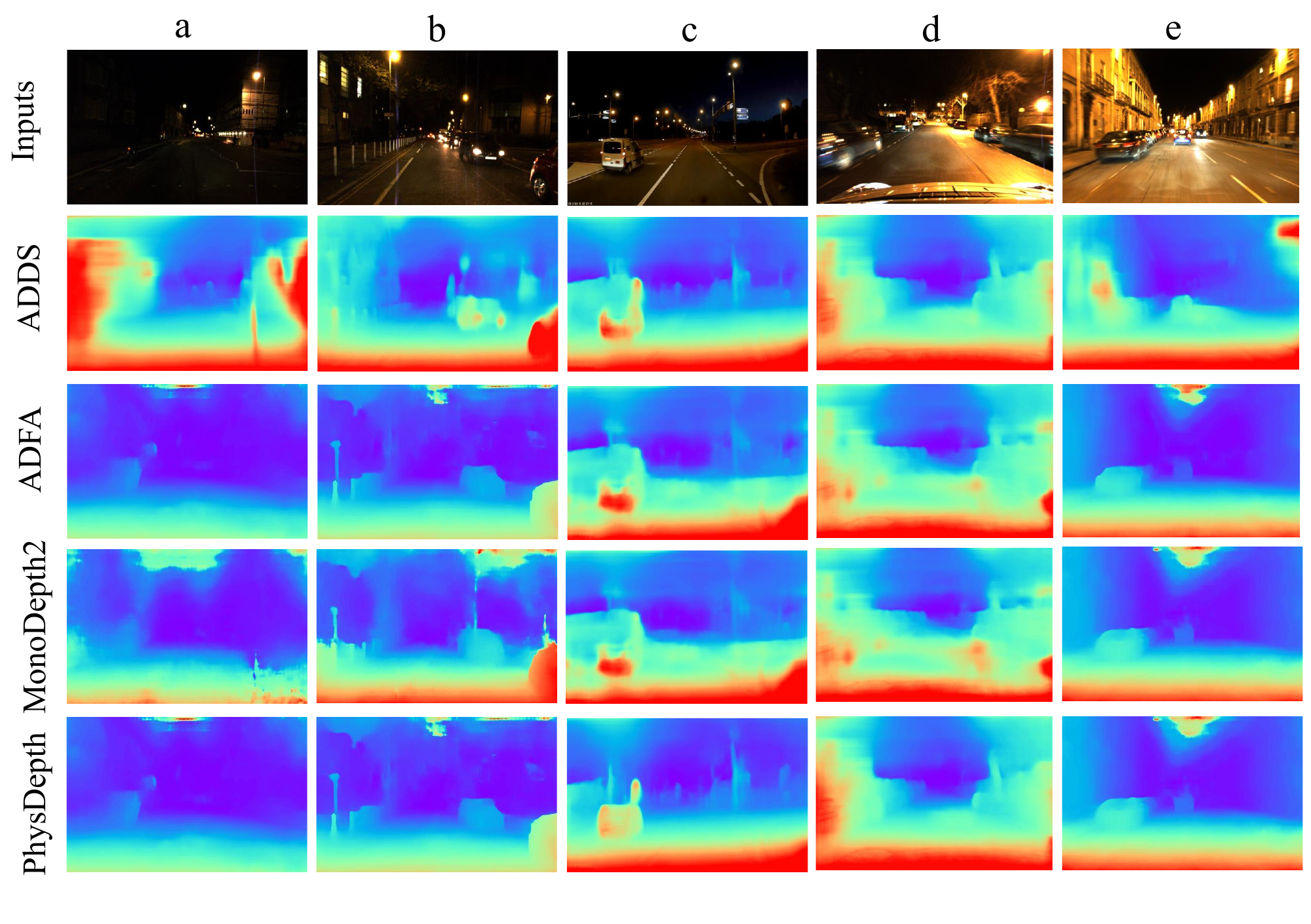}
        \vspace{-6mm}
	\caption{\textbf{Qualitative Results - RobotCar-Night:} Depth estimated on five different test images (row: Inputs) using our model (row: PhysDepth) and three SOTA (remaining rows) for comparison.}
	\label{OxforRobot-Night visual}
    \vspace{-4mm}
\end{figure*}

\subsection{Red Channel Attenuation Loss (RCA loss)}
\label{sec:rca loss}

To guide the training processes of PPM, we introduce RCA loss (Figure \ref{fig:network} - c). 
RCA loss is based on Beer-Lambert law (Equation \ref{eq:Beer-Lambert law}) and Rayleigh scattering theory (Equation \ref{eq:Rayleigh scattering}).

\begin{equation}
    \label{eq:Beer-Lambert law}
    I = I_0 e^{- \mu d_{R}}
\end{equation}

Specifically, in our model, we consider Equation \ref{eq:Beer-Lambert law} for each pixel, where $ I_0 $ denotes the intensity of the luminous point. 
$ I $ denotes the light intensity after attenuation.
$ \mu $ is a constant and $ d $ is depth value. $ e^{- \mu d_{R}} $ is the attenuation item. 
Also, we discuss the different items in Section \ref{sec:Additional Quantitative Results} in Supplementary Material.

Equation \ref{eq:Beer-Lambert law} suggests the possibility of deriving the depth value, denoted as $ d_{R} $ from the information contained in $ I $ and $ I_0 $. This insight motivates us to employ it as a loss function during the training of PPM. To facilitate this, we rephrase Equation \ref{eq:Beer-Lambert law} to Equation \ref{eq:rca loss} as:

\begin{equation}
    \label{eq:rca loss}
    d_{R} = -\frac{1}{\mu} \ln I+ \frac{\ln I_0}{\mu} 
\end{equation}

We then use gradient Spherical Gaussian (SG) \cite{li2022physically, wang2009all, wang2021learning} to model $ I_0 $ in Equation \ref{eq:sg}, where 
$ \lambda \in \mathbb{R} $ is a learnable parameter.
It is because the naive application of the Beer-Lambert law would require solving for the per-pixel intrinsic brightness $I_0$, which is ill-posed from a single image. 
We instead simplify this term by modeling its contribution as a single, globally learnable parameter $\lambda$.
We are therefore not claiming to model the true $I_0$ of every pixel. 
Rather, the term $\frac{1}{\mu}(g\lambda-1)$ functions as a learned scalar bias in the regression. This simplification makes the problem tractable and prevents the model from learning a trivial identity by encoding depth in the $I_0$ term.

In our experiments, we empirically set $ g = 1.3938 $ for the best accuracy.

\begin{equation}
    \label{eq:sg}
    I_0 =  e ^ {g \lambda - 1}
\end{equation}

Additionally, as discussed in Section \ref{sec:PPM}, the phenomenon of Rayleigh scattering theory drives our decision to focus on the attenuation of red channel in our architecture. 
Thus, we replace the RGB image $ I $ with red channel image $ R $, and $ f(R) $ represents the feature from the last layer of PPM.
Mathematically, our goal is for the final layer to learn an identity map
ping, $f(R) = R$.
And we apply the law to $f(R)$, the output of the PPM, rather than the raw pixel intensity $R$. 
This is a principled design choice, as raw pixel values are a poor proxy for true physical radiance. The $R$ value is corrupted by non-linearities from the camera's Image Signal Processor (e.g., gamma correction, tonemapping) and sensor noise.
The PPM ($f(R)$) is trained to learn a linearized, denoised representation of the scene's radiance.
Therefore, $f(R)$ serves as a more suitable and physically valid input for the Beer-Lambert law than $R$ itself Table \ref{tab:r and fr} in supplemental material. 
Taking gradient SG and $ f(R) $ into account, we can now reformulate Equation \ref{eq:rca loss} to Equation \ref{eq:rca loss r and c} as follow:

\begin{equation}
    \label{eq:rca loss r and c}
    d_R = -\frac{1}{\mu} \ln f(R) + \frac{1}{\mu} (g\lambda - 1)
\end{equation}

Then we apply Equation \ref{eq:rca loss r and c} to train PPM. Specifically, when we know $ d_R $, we use $ \mathcal{L}_2 $ to denote L2 loss and compute $ \mathcal{L}_2 $ between $ d_R $ and $ d_\text{ES} $. $ d_\text{ES} $ is the estimated depth map from Base MDE decoder.
\textit{Because the attenuation has a functional relation $ \ln f(R) $ with depth value $ d_{R} $, it can be used to supervise the training of PPM}, which is the mathematical foundation of our design choice.

\textbf{Total Training Loss Function, }

RCA loss serves as a self-supervised signal to guide the training processes of PPM. 
Besides RCA loss, we also train our model with another supervisory signal: photometric reconstruction loss function $ \mathcal{L}_\text{p} $ from SfMLearner \cite{zhou2017unsupervised}. It requires four inputs as Figure \ref{fig:network} shows: Image $ R_t, R_{t-1} $, Transformation $ T_{t \rightarrow t-1} $, predicted depth map $ D_t $, and camera intrinsics matrix $ K $.
We add the weak velocity supervision $\mathcal{L}_\text{v}$ \cite{guizilini20203d} to address 
scale ambiguity and ensure consistent predictions of photometric reconstruction \cite{gasperini2023robust}.
Also following \cite{godard2019digging,gasperini2023robust}, we consider partial occlusions via the minimum reprojection error $\mathcal{L}_\text{p}$, and we ignore static pixels. 
Another loss $ \mathcal{L}_\text{s} $ is used to promote smoothness and preserves edges \cite{godard2017unsupervised}
The total loss function is $ \mathcal{L} = \mathcal{L}_\text{p} + \mathcal{L}_2 + \mathcal{L}_\text{v} + \mathcal{L}_\text{s}$.

\begin{figure*}[t]
	\centering
	\includegraphics[width=\linewidth]{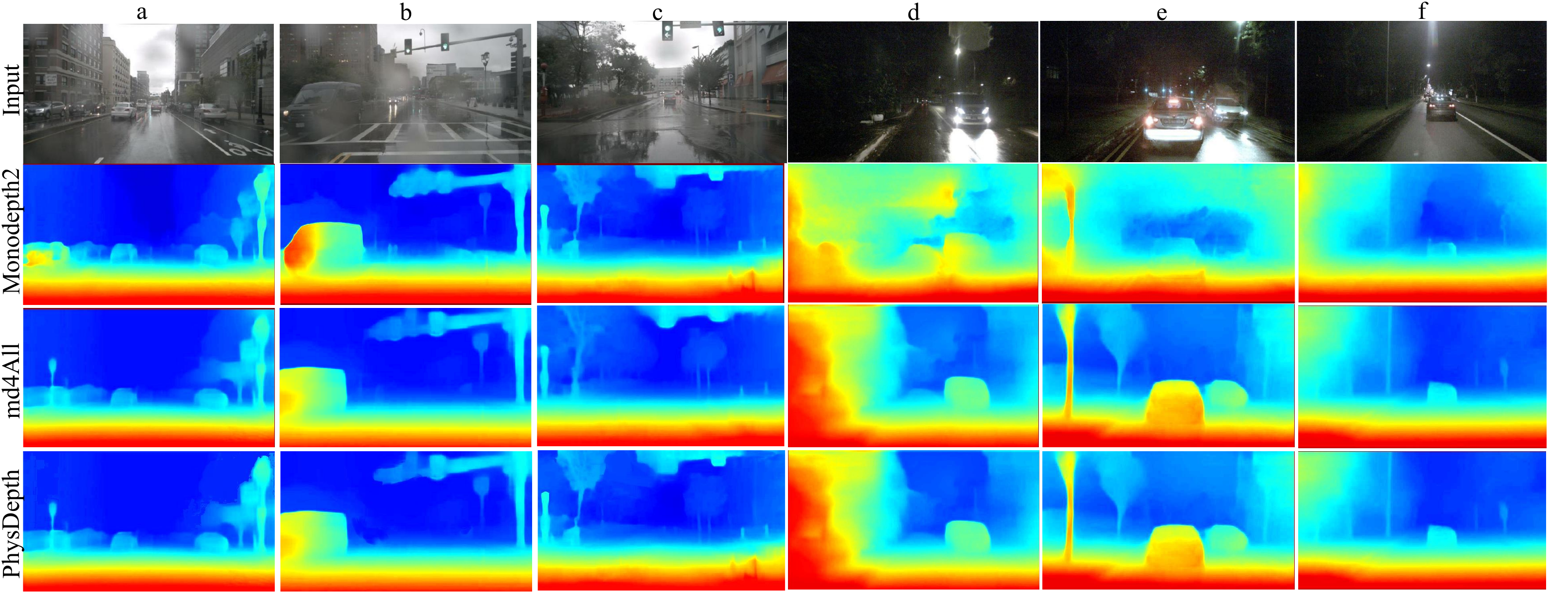}
        \vspace{-7mm}
	\caption{\textbf{Qualitative Results - nuScence-Rain and nuScence-Night:} Depth estimated on different test images using PhysDepth and two SOTAs for comparison.}
	\label{nuSence visual}
    \vspace{-2mm}
\end{figure*}

\begin{table}[ht]
    \centering
    \footnotesize
    \caption{\textbf{Quantitative Results - RobotCar-Night  \cite{RobotCarDatasetIJRR}:} We scale depth up to 50 meters as same as  \cite{gasperini2023robust,vankadari2023sun}. The best results are in \textbf{bold} for each metric. The results presented here are without post-processing. AD and DD mean different training strategies for md4All.}
    \vspace{-2mm}
    \begin{tabular}{l|cc|c}
    \toprule
    \multicolumn{1}{c|}{Methods} & \multicolumn{1}{c|}{Abs Rel}        & \multicolumn{1}{c|}{RMSE}           & \multicolumn{1}{c}{$\delta<1.25$} \\
    \hline
    \textbf{} & \multicolumn{2}{c|}{lower's better} & \multicolumn{1}{c}{higher's better} \\
    \hline
    
    MonoDepth2 \cite{godard2019digging}                   & \multicolumn{1}{c|}{0.453}          & \multicolumn{1}{c|}{11.420}          & \multicolumn{1}{c}{0.700} \\ \hline
    DeFeat-Net \cite{spencer2020defeat}                   & \multicolumn{1}{c|}{0.334}          & \multicolumn{1}{c|}{8.606}          & \multicolumn{1}{c}{0.586} \\ \hline
    
    ADDS \cite{liu2021self}                         & \multicolumn{1}{c|}{0.287}          & \multicolumn{1}{c|}{7.985}         & \multicolumn{1}{c}{0.490} \\ \hline
    
    DepthSegNet24 \cite{thanh2024depthsegnet24}                         & \multicolumn{1}{c|}{0.195}          & \multicolumn{1}{c|}{ 5.936}          & \multicolumn{1}{c}{0.734} \\
    \hline
    
    ADFA \cite{vankadari2020unsupervised}                         & \multicolumn{1}{c|}{0.231}          & \multicolumn{1}{c|}{8.800}          & \multicolumn{1}{c}{0.620} \\
    \hline
    
    RNW \cite{wang2021regularizing}                          & \multicolumn{1}{c|}{0.185}          & \multicolumn{1}{c|}{{{6.549}}}    & \multicolumn{1}{c}{0.733} \\ \hline
    
    WSGD \cite{vankadari2023sun}    & \multicolumn{1}{c|}{0.174}    & \multicolumn{1}{c|}{6.302}          & \multicolumn{1}{c}{0.754} \\ \hline
    
    STEPS \cite{zheng2023steps} & \multicolumn{1}{c|}{0.170}    & \multicolumn{1}{c|}{ 6.797}          & \multicolumn{1}{c}{0.758} \\ \hline
    
    DepthDark \cite{zeng2025depthdark}   & \multicolumn{1}{c|}{0.157}    & \multicolumn{1}{c|}{4.284}          & \multicolumn{1}{c}{0.760} \\ \hline
    
    md4All-AD \cite{gasperini2023robust}                      & \multicolumn{1}{c|}{0.124}    & \multicolumn{1}{c|}{3.612}          & \multicolumn{1}{c}{0.858} \\ \hline
    
    md4All-DD \cite{gasperini2023robust}                      & \multicolumn{1}{c|}{0.122}    & \multicolumn{1}{c|}{3.604}          & \multicolumn{1}{c}{0.849} \\
    \hline
    
    {Syn2Real-Depth} \cite{yan2025synthetic}                         & \multicolumn{1}{c|}{0.110} & \multicolumn{1}{c|}{3.455} & \multicolumn{1}{c}{0.862} \\
    \hline
     
    \textbf{{PhysDepth}}                         & \multicolumn{1}{c|}{\cellcolor{db}\textbf{0.109}} & \multicolumn{1}{c|}{\cellcolor{db}\textbf{3.450}} & \multicolumn{1}{c}{\cellcolor{db}\textbf{0.863}} \\
    
    \bottomrule
    
    \end{tabular}
    \label{OxforRobot-Night qua}
\end{table}
   
\begin{table}[t]
    \footnotesize
    \centering
    \caption{\textbf{Quantitative Results - nuScenes-Night \cite{caesar2020nuscenes}:} We scale depth up to 80 meters as same as \cite{gasperini2021r4dyn,gasperini2023robust}. The best results are in \textbf{bold} for each metric. All results presented here are without post-processing.  AD and DD mean different training strategies for md4All.}
    \vspace{-2mm}
    \begin{tabular}{l|cc|c}
    \toprule
    \multicolumn{1}{c|}{Methods} & \multicolumn{1}{c|}{Abs Rel}        & \multicolumn{1}{c|}{RMSE}           & \multicolumn{1}{c}{$\delta<1.25$} \\
    \hline
    \textbf{} & \multicolumn{2}{c|}{lower's better} & \multicolumn{1}{c}{higher's better} \\
    \hline
    
    STEPS \cite{zheng2023steps} & \multicolumn{1}{c|}{0.474}    & \multicolumn{1}{c|}{ 13.158}          & \multicolumn{1}{c}{0.334} \\ \hline
    
    SRNSD \cite{cong2024srnsd}    & \multicolumn{1}{c|}{0.352}    & \multicolumn{1}{c|}{11.942}          & \multicolumn{1}{c}{0.419} \\ \hline
    Robust-Depth \cite{saunders2023self} & \multicolumn{1}{c|}{0.276}   & \multicolumn{1}{c|}{10.470}  & \multicolumn{1}{c}{0.607} \\
    \hline
    
    RNW \cite{wang2021regularizing}                             & \multicolumn{1}{c|}{0.290}   & \multicolumn{1}{c|}{11.289} & \multicolumn{1}{c}{0.557} \\
    \hline
    
    R4Dyn \cite{gasperini2021r4dyn}                                    & \multicolumn{1}{c|}{0.219}   & \multicolumn{1}{c|}{ 10.542}       & \multicolumn{1}{c}{0.623} \\
    \hline
    
    DepthDark \cite{zeng2025depthdark}   & \multicolumn{1}{c|}{0.210}    & \multicolumn{1}{c|}{7.764}          & \multicolumn{1}{c}{0.630} \\ \hline
    
    Syn2Real-Depth \cite{yan2025synthetic}                                  & \multicolumn{1}{c|}{0.171}   & \multicolumn{1}{c|}{7.689}  & \multicolumn{1}{c}{0.732} \\
    \hline
    
    MonoDepth2 \cite{godard2019digging}                           & \multicolumn{1}{c|}{0.147}   & \multicolumn{1}{c|}{6.941}  & \multicolumn{1}{c}{0.827} \\
    \hline
   
    md4All-AD \cite{gasperini2023robust}                                    & \multicolumn{1}{c|}{0.182}   & \multicolumn{1}{c|}{6.372}  & \multicolumn{1}{c}{0.753} \\
    \hline
    
    md4All-DD \cite{gasperini2023robust}                                    & \multicolumn{1}{c|}{0.128}   & \multicolumn{1}{c|}{6.607}  & \multicolumn{1}{c}{0.845} \\
    \hline
    
    \textbf{PhysDepth}                                    & \multicolumn{1}{c|}{\cellcolor{db}\textbf{0.118}}   & \multicolumn{1}{c|}{\cellcolor{db}\textbf{6.349}}  & \multicolumn{1}{c}{\cellcolor{db}\textbf{0.853}} \\
    \bottomrule
    \end{tabular}
    \label{tab:nuScenes}
\end{table}

\begin{table}[ht]
    \centering
    \footnotesize
    \caption{\textbf{Quantitative Results - nuScenes-Rain \cite{caesar2020nuscenes}:} We scale depth up to 80 meters as same as \cite{gasperini2021r4dyn,gasperini2023robust}. The best results are in \textbf{bold} for each metric. All results presented here are without post-processing. AD and DD mean different training strategies for md4All.}
    \vspace{-2mm}
    \begin{tabular}{l|cc|c}
    \toprule
    \multicolumn{1}{c|}{Methods} & \multicolumn{1}{c|}{Abs Rel} & \multicolumn{1}{c|}{RMSE} & \multicolumn{1}{c}{$\delta<1.25$} \\
    \hline
    \textbf{} & \multicolumn{2}{c|}{lower's better} & \multicolumn{1}{c}{higher's better} \\
    \hline
    
    RNW \cite{vankadari2020unsupervised} & \multicolumn{1}{c|}{0.295} & \multicolumn{1}{c|}{9.341} & \multicolumn{1}{c}{0.572} \\ \hline

    R4Dyn \cite{gasperini2021r4dyn} & \multicolumn{1}{c|}{0.243} & \multicolumn{1}{c|}{10.055} & \multicolumn{1}{c}{0.570} \\ \hline

    AdaBins \cite{bhat2021adabins} & \multicolumn{1}{c|}{0.234} & \multicolumn{1}{c|}{7.088} & \multicolumn{1}{c}{0.616} \\ \hline
    
    md4All-DD \cite{gasperini2023robust} & \multicolumn{1}{c|}{0.214} & \multicolumn{1}{c|}{8.376} & \multicolumn{1}{c}{0.680} \\ \hline

     MonoDepth \cite{liu2021self} & \multicolumn{1}{c|}{0.157} & \multicolumn{1}{c|}{7.453} & \multicolumn{1}{c}{0.795} \\ \hline

     md4All-AD \cite{gasperini2023robust} & \multicolumn{1}{c|}{0.141} & \multicolumn{1}{c|}{6.915} & \multicolumn{1}{c}{{0.830}} \\ \hline

    Syn2Real-Depth \cite{yan2025synthetic} & \multicolumn{1}{c|}{{0.133}} & \multicolumn{1}{c|}{6.926} & \multicolumn{1}{c}{0.817} \\ \hline
        
    \textbf{PhysDepth} & \multicolumn{1}{c|}{\cellcolor{db}\textbf{0.130}} & \multicolumn{1}{c|}{\cellcolor{db}\textbf{6.910}} & \multicolumn{1}{c}{\cellcolor{db}\textbf{0.835}} \\
    
    \bottomrule
    
    \end{tabular}
    \label{tab:nuScence rain}
\end{table}

\section{Experiments}

\textbf{Implementation Details:}
Our model is built using PyTorch \cite{paszke2019pytorch}.
We train our model with a batch size of 4, Adam optimizer \cite{kingma2014adam} is used with $\beta_{1} = 0.9$ and $\beta_{2} = 0.999$, we set the learning rate to $0.0002$.
Meanwhile, we follow Zhou et al. \cite{zhou2017unsupervised}, which we multiply the predicted depth maps by a scalar $\hat{s}$ to match the median with the groundtruth, i.e., $\hat{s} = \operatorname{median}(D_\text{gt})/\operatorname{median}(D_\text{pred})$.

\textbf{Datasets:}
We evaluate PhysDepth on three datasets (RobotCar-Night \cite{RobotCarDatasetIJRR}, nuScenes-Night \cite{caesar2020nuscenes}, and nuScenes-Rain \cite{caesar2020nuscenes})

to demonstrate that PhysDepth is accurate and robust under challenging environments.
Our model is trained on those three datasets, respectively. 

For baselines, we use any provided training weights from the original authors.

Afterwards, these weights are fine-tuned on all datasets for a fair comparison. 
More details about the training and testing on the datasets can be found in Section \ref{sec:suppl DataSet Details} of the Supplementary Material.

We evaluate our proposed model both quantitatively and qualitatively.
For quantitative analysis, we perform the comparison against SOTA methods on all three datasets mentioned above.
We use three metrics, same as those used in \cite{eigen2014depth, godard2019digging} - Abs Rel, RMSE, and $\delta < 1.25$.
For qualitative analysis, we compare estimated depth maps from PhysDepth and SOTA models using images from RobotCar-Night and nuScenes (night and rain).

\subsection{Quantitative Results}
\label{exp:quant_Results}

Tables \ref{OxforRobot-Night qua}, \ref{tab:nuScenes} show the quantitative results on the Oxford RobotCar-Nigh, and nuScenes-Night, respectively. 
As Table \ref{OxforRobot-Night qua} and \ref{tab:nuScenes} show, our model outperforms all SOTA models across all metrics, emphasizing its overall robustness and accuracy. 

This could be attributed to the Physical Prior module and RCA loss, modeling the physical priors of Rayleigh scattering theory and Beer-Lambert law. 

These physical prior are more effective than other previously adopted domain-specific features in improving MDE (ADDS \cite{liu2021self}, and ADFA \cite{vankadari2020unsupervised}). 

Among Tables \ref{OxforRobot-Night qua}, \ref{tab:nuScenes}, md4All \cite{gasperini2023robust} and Syn2Real-Depth \cite{yan2025synthetic}  are two strong baselines, which both adopt GAN-based model and odometry information to achieve better performance.

WSGD \cite{vankadari2023sun}  uses three techniques to allow the existing photometric loss to work for both day and challenging condition images.

However, either photometric loss or additional constraints still ignore the light attenuation, leading to suboptimal results.
Meanwhile, STEPS \cite{zheng2023steps} learns nighttime condition image enhancer to improve nighttime MDE. 

R4Dyn \cite{gasperini2021r4dyn}'s good performance mainly comes from weak radar supervision.
However, PhysDepth outperformed R4Dyn because PhysDepth learns robust physical features while also using weak velocity supervision. 
RNW \cite{wang2021regularizing} is another strong baseline, which performs better than other SOTA by using image enhancement module to enhance image visibility and contrast. 

Table \ref{tab:nuScence rain} provides a quantitative evaluation of our model's performance on the challenging nuScenes-Rain dataset, demonstrating its robustness to adverse weather conditions.
We compare PhysDepth against several SOTA baselines. 
The results highlight that PhysDepth achieves the best-performing RMSE score of 6.910, Abs Rel of 0.130, $\delta < 1.25$ of 0.835, significantly outperforming competitors. 

Overall, PhysDepth's high accuracy is attributed to \textit{ physical priors: light attenuation for red channel images, which is more robust and effective in improving depth estimation accuracy}. 

\subsection{Qualitative Results}
\label{exp:quali_Results}

Figure \ref{OxforRobot-Night visual} 
shows results on the RobotCar-Night dataset, qualitatively demonstrating that PhysDepth provides more accurate MDE results under challenging environments. 
While our depth map clearly delineates the car, ADDS fails to capture its boundary; ADAF and MonoDepth2 detect the car, but their outputs show blurred car boundaries. This is because PhysDepth depends on the red channel, which preserves more complete texture information and minimizes noise from other channels under extremely low-lighting conditions. Moreover, in Figure \ref{OxforRobot-Night visual} - c, PhysDepth effectively captures the car’s boundary while other SOTA models all fail to estimate the car's back windshield area due to low-light and reflection, showcasing PhysDepth's robustness under challenging environments. 

Figure \ref{nuSence visual} presents a qualitative comparison of our PhysDepth model against SOTA methods Monodepth2 and md4All on challenging test images from the nuScenes-Night and nuScenes-Rain datasets. 
The top row (Input) displays a variety of challenging conditions, including daytime rain with reflections (a to c) and extreme nighttime glare from headlights (d to f). 
The subsequent rows visually demonstrate the predicted depth maps from each model, highlighting our method's ability to produce coherent and stable geometric estimates compared to the baselines under these adverse scenarios.

\subsection{Plugin Analysis}
\label{exp:plug_Results}
As discussed before, \textit{PPM and Loss Functions (RCA loss, Weak Velocity Loss, and Photometric Loss)} together serve as a plugin to enhance the accuracy of existing MDE models. 
Table \ref{tab:plug-in} demonstrates the effectiveness and generality of PhysDepth. 
We plug in our modules (PPM + Loss Functions) to six SOTA backbones.
The results show consistent performance improvements over the baselines on all three datasets across all metrics.

For instance, md4All-AD improves from 0.182 Abs Rel to 0.179 on nuScenes-Night when our module is added. MonoDepth enhanced from 0.453 Abs Rel to 0.448 on RobotCar-Night when plugged in.
This robustness generalizes to the nuScenes-Rain dataset, where PhysDepth + RNW (3.791 RMSE) dramatically outperforms the RNW baseline (9.341 RMSE), and PhysDepth + md4All-DD (0.210 Abs Rel) improves over its baseline (0.214 Abs Rel). 

The results show that our framework provides a \textbf{model-agnostic enhancement} that generalizes from darkness to other challenging low-visibility environments, such as rain.

\begin{table*}[ht]
    \footnotesize
    \centering
    \caption{\textbf{Plug-and-Play Validation and Generalization to Adverse Weather.} 
    We apply our PhysDepth framework as a plug-in module to several existing SOTA models. 
    The results show a consistent performance improvement on RobotCar-Night, nuScenes-Night, and nuScenes-Rain, demonstrating the effectiveness, 'plug-and-play' capability, and robustness of our approach against challenging conditions beyond darkness.}
    \label{tab:plug-in}
    
    \begin{tabular}{l|c|c|c|c|c|c|c|c|c}
    \toprule
    \multicolumn{1}{c|}{Methods}     & \multicolumn{3}{c|}{RobotCar-Night} & \multicolumn{3}{c|}{nuScenes-Night} & \multicolumn{3}{c}{nuScenes-Rain} \\ 
    \cline{2-10}
    & Abs Rel & RMSE & $\delta < 1.25$ & Abs Rel & RMSE & $\delta < 1.25$ & Abs Rel & RMSE & $\delta < 1.25$ \\ 
    \hline
    \textbf{} & \multicolumn{2}{c|}{lower's better} & \multicolumn{1}{c|}{higher's better} & \multicolumn{2}{c|}{lower's better} & \multicolumn{1}{c|}{higher's better} & \multicolumn{2}{c|}{lower's better} & \multicolumn{1}{c}{higher's better} \\
    \hline
    
    RNW \cite{vankadari2020unsupervised} & 0.185 & 6.549 & 0.733 & 0.290 & 11.289 & 0.557 & 0.295 &  9.341 & 0.572 \\ 
    PhysDepth + RNW  & \cellcolor{db}0.180 & \cellcolor{db}6.531 & \cellcolor{db}0.739& \cellcolor{db}0.286 & \cellcolor{db}11.280 & \cellcolor{db}0.562 &  \cellcolor{db}0.290  & \cellcolor{db}3.791  & \cellcolor{db}0.584 \\ 
    \hline
    Syn2Real-Depth \cite{yan2025synthetic} & 0.110 & 3.455 & 0.862 & 0.171 & 7.689 & 0.732 & 0.133 & 6.926 &0.817 \\ 
    PhysDepth + S2R \cite{yan2025synthetic}  & \cellcolor{db}0.105 & \cellcolor{db}3.450 & \cellcolor{db}0.869 & \cellcolor{db}0.166 & \cellcolor{db}7.681 & \cellcolor{db}0.737 & \cellcolor{db}0.129 & \cellcolor{db}6.921 & \cellcolor{db}0.821 \\ 
    \hline
    MonoDepth \cite{liu2021self} & 0.453 & 11.420 & 0.700& 0.147 & 6.941 & 0.827 & 0.157 & 7.453 & 0.795 \\ 
    PhysDepth + MonoDepth \cite{liu2021self}  & \cellcolor{db}0.448 & \cellcolor{db}11.413 & \cellcolor{db}0.708 & \cellcolor{db}0.143& \cellcolor{db}6.935& \cellcolor{db}0.823 & \cellcolor{db}0.152 & \cellcolor{db}7.447 & \cellcolor{db}0.799 \\ 
    \hline
    md4All-AD \cite{gasperini2023robust} &  0.124 & 3.612 & 0.858  &  0.182 & 6.372 & 0.753  & 0.141 & 6.915 & 0.830\\ 
    PhysDepth + md4All-AD\cite{gasperini2023robust} &  \cellcolor{db}0.119 & \cellcolor{db}3.607  & \cellcolor{db}0.862  & \cellcolor{db}0.179  & \cellcolor{db}6.367  & \cellcolor{db}0.758  & \cellcolor{db}0.137  & \cellcolor{db}6.910  & \cellcolor{db}0.836 \\ 
    \hline 
    md4All-DD \cite{gasperini2023robust} & 0.122  & 3.604  & 0.849 & 0.128  & 6.607  & 0.845  & 0.214 & 8.376 & 0.680\\ 
    PhysDepth + md4All-DD\cite{gasperini2023robust} & \cellcolor{db}0.117  & \cellcolor{db}3.600  & \cellcolor{db}0.853  &  \cellcolor{db}0.125 &  \cellcolor{db}6.601 & \cellcolor{db}0.849  &  \cellcolor{db}0.210 &  \cellcolor{db}8.371 & \cellcolor{db}0.687 \\ 
    \hline
    DepthDark \cite{zeng2025depthdark} & 0.157 & 4.284 & 0.760 & 0.210 & 7.689 & 0.630 & \multirow{2}{*}{-} & \multirow{2}{*}{-} & \multirow{2}{*}{-} \\ 
    PhysDepth+DepthDark \cite{zeng2025depthdark}  & \cellcolor{db}0.153 & \cellcolor{db}4.280& \cellcolor{db}0.764& \cellcolor{db}0.206&\cellcolor{db}7.683 & \cellcolor{db}0.635 & & & \\
    \bottomrule
    \end{tabular}
    \end{table*}

\section{Ablation Study} 
\label{sec:ablation}

Table \ref{tab:OxforRobot-Night abl} presents our core Ablation Study on the RobotCar-Night dataset.
We tested four different base architectures: 
convNeXt-T, MiDaS, ViT-B/16, and ViT-B/32.
A clear and consistent trend emerged across all backbones: the baseline models (e.g., ViT-B/16 baseline at 0.678 Abs Rel) performed poorly; adding only the PPM module provided a significant boost (e.g., ViT-B/16+PPM at 0.318 Abs Rel); and finally, adding both the PPM+RCA loss yielded the full, substantial performance gain (e.g., ViT-B/16+PPM+RCA at 0.117 Abs Rel).
This strongly demonstrates that both modules are essential and effective, with the ViT-B/32+PPM+RCA combination achieving the overall best results across all metrics (0.114 Abs Rel, 3.522 RMSE, and 0.863 $\delta < 1.25$).

\begin{table}
    \footnotesize
    \centering
    \caption{\textbf{Ablation Study - RobotCar-Night \cite{wang2021regularizing}:} We evaluate different backbone architectures and module combinations. The best results are in \textbf{bold} for each metric.}
    \vspace{-3mm}
    \begin{tabular}{l|cc|c}
    \toprule
    \multicolumn{1}{c|}{Method}  & \multicolumn{1}{c|}{Abs Rel} & \multicolumn{1}{c|}{RMSE} & \multicolumn{1}{c}{$\delta < 1.25$} \\
    \hline
    \textbf{} & \multicolumn{2}{c|}{lower's better} & \multicolumn{1}{c}{higher's better} \\
    \hline
    
    convNeXt-T & 0.690 & 7.250 & 0.640 \\
    convNeXt-T+PPM & 0.330 & 4.450 & 0.665 \\
    convNeXt-T+PPM+RCA & 0.120 & 3.550 & 0.850 \\
    \hline
    
    MiDaS & 0.671 & 7.098 & 0.651 \\
    MiDaS+PPM & 0.311 & 4.308 & 0.672 \\
    MiDaS+PPM+RCA & 0.118 & 3.529 & 0.858 \\
    \hline
    
    ViT-B/16 & 0.678 & 7.110 & 0.651 \\
    ViT-B/16-T+PPM & 0.318 & 4.315 & 0.675 \\
    \cellcolor{yellow!50}\textbf{ViT-B/16+PPM+RCA} & \cellcolor{yellow!50}\textbf{0.117} & \cellcolor{yellow!50}\textbf{3.525} & \cellcolor{yellow!50}\textbf{0.860} \\
    \hline
    
    ViT-B/32 & 0.675 & 7.105 & 0.655 \\
    ViT-B/32+PPM & 0.314 & 4.310 & 0.678 \\
    \cellcolor{db}\textbf{ViT-B/32+PPM+RCA} & \cellcolor{db}\textbf{0.114} & \cellcolor{db}\textbf{3.522} & \cellcolor{db}\textbf{0.863} \\
    \bottomrule
    
    \end{tabular}
    \label{tab:OxforRobot-Night abl}
\end{table}

\begin{table}[ht]
    \centering
    \footnotesize
    \caption{\textbf{Ablation Study of Using Different Channel Images on RobotCar-Night Dataset \cite{RobotCarDatasetIJRR}:} R means Red Channel, so do B (Blue Channel) and G (Green Channel), RGB. The best results are in \textbf{bold}.}    \vspace{-2mm}
    \begin{tabular}{l|cc}
    \toprule
    \multicolumn{1}{c|}{Methods} & \multicolumn{1}{c|}{Abs Rel} & \multicolumn{1}{c}{RMSE} \\
    \hline
    \textbf{} & \multicolumn{2}{c}{lower's better} \\
    \hline
    
    PhysDepth+RGB & \multicolumn{1}{c|}{0.171} & \multicolumn{1}{c}{4.028} \\ \hline
    
    PhysDepth+B+R & \multicolumn{1}{c|}{0.155} & \multicolumn{1}{c}{3.921} \\ \hline
    
    PhysDepth+G+R & \multicolumn{1}{c|}{0.156} & \multicolumn{1}{c}{3.897} \\ \hline
    
    PhysDepth+B+G & \multicolumn{1}{c|}{0.159} & \multicolumn{1}{c}{3.842} \\ \hline
    
    PhysDepth+B & \multicolumn{1}{c|}{0.120} & \multicolumn{1}{c}{3.540} \\ \hline
    
    PhysDepth+G & \multicolumn{1}{c|}{0.119} & \multicolumn{1}{c}{3.531} \\ \hline
    
    \textbf{PhysDepth+R} & \multicolumn{1}{c|}{\cellcolor{db}\textbf{0.117}} & \multicolumn{1}{c}{\cellcolor{db}\textbf{3.525}} \\
    
    \bottomrule
    
    \end{tabular}
    \label{tab:abl_channel}
    \end{table}
    
In Table \ref{tab:abl_channel}, our model is trained on channel-specific images from the RobotCar-Night dataset. Those channels are: red channel(R), blue channel(B), green channel(G), and three-channels (RGB).
As shown in Table \ref{tab:abl_channel}, the PhysDepth+R only uses the red channel, which outperforms all other channel/channel-combinations, validating the foundations of our model design: 1) \textit{using red channel images as input}; 2) \textit{adopting physical-prior-knowledge to enhance the accuracy and robustness of MDE}. 

Particularly, the findings in Table \ref{tab:abl_channel} and in Table \ref{tab:abl_channel_RobotCar_Night_baseline} are consistent.
We can observe that using single-channel image settings (R, G, or B) performs better than using two-channel image (R+G, R+B, or G+B). While using two-channel image settings outperforms using three-channel (RGB) images. 
This can be attributed to the fact that: \textit{combining multiple channels could introduce noise, especially in low-light conditions} (e.g., Green and Blue channels tend to lose more texture information due to Rayleigh Scattering theory).
That is, the additional channels (Green and Blue), instead of contributing significantly useful image information, could "noise" the primary channel (Red) and confuse depth estimation models, leading to compromised MDE results.
Additionally, the red channel performs best among all single-channel settings, followed by the green channel; the blue channel performs the worst.
The results yield with Rayleigh scattering theory (Equation \ref{eq:Rayleigh scattering}) that $I_\text{red} < I_\text{green} < I_\text{blue}$, which $ I $ represents intensity of scattering. 

\begin{table}[ht]
    \centering
    \footnotesize
    \caption{\textbf{Ablation Study of Attenuation Item:} Using red channel images from the RobotCar-Night dataset \cite{wang2021regularizing} and the depth range is up to 80m. Linear means that we treat light attenuation as a linear function. And so do Quadratic and Exponential. The best results are in \textbf{bold}.}    
    \begin{tabular}{l|cc}
    \toprule
    \multicolumn{1}{c|}{Methods} & \multicolumn{1}{c|}{Abs Rel} & \multicolumn{1}{c}{RMSE} \\
    \hline
    \textbf{} & \multicolumn{2}{c}{lower's better} \\
    \hline
    
    PhysDepth+Linear & \multicolumn{1}{c|}{0.122} & \multicolumn{1}{c}{3.741} \\ \hline
    
    PhysDepth+Quadratic & \multicolumn{1}{c|}{0.129} & \multicolumn{1}{c}{3.735} \\ \hline
    
    \textbf{PhysDepth+Exponential} & \multicolumn{1}{c|}{\cellcolor{db}\textbf{0.121}} & \multicolumn{1}{c}{\cellcolor{db}\textbf{3.657}} \\
    
    \bottomrule
    
    \end{tabular}
    \vspace{-3mm}
    \label{tab:attenuation function}
    \end{table}

We also conduct a study of attenuation item $e^{- \mu d_R}$ (Exponential) in Table \ref{tab:attenuation function}. While the other parts in Equation \ref{eq:Beer-Lambert law} remain the same, we first change $e^{- \mu d_R}$ to linear function (Equation \ref{eq:linear}). And then, we change $e^{- \mu d_R}$ to quadratic function (Equation \ref{eq:quadratic}). 
For the experiment settings, we set $a=0.0095, b=0.05$ for Equation \ref{eq:linear} \cite{attenuation}. We then set $a=0.0095, b=0.05, c= 1.0$ for Equation \ref{eq:quadratic} \cite{attenuation}. All other experiment settings stay the same as in Table \ref{OxforRobot-Night qua}.
The results demonstrate our design choice of using $e^{- \mu d_R}$ as attenuation item.

The supplementary material includes additional results and discussions, such as an ablation study on various light attenuation functions, analyses of different depth ranges, additional qualitative results, failure cases, etc.

\section{Conclusion and Limitations} 
\label{sec:conclusion}
In this work, we identify a critical weakness in modern monocular depth estimation, and demonstrate that SOTA models often fail in challenging environments by overlooking physical information. We substantiate this claim through an empirical study. 
We propose PhysDepth, a novel, plug-and-play framework designed to infuse physical priors into existing MDE backbones. 

Extensive quantitative and qualitative evaluations on the RobotCar-Night, nuScenes-Night, and nuScenes-Rain datasets demonstrate that PhysDepth achieves SOTA accuracy. Furthermore, we validate the generality and model-agnostic nature of our approach. By applying PhysDepth as a plugin to multiple existing SOTA backbones, we observe consistent and significant performance improvements, confirming its utility as a plug-and-play solution.

While PhysDepth demonstrates remarkable robustness, we acknowledge limitations in extreme scenarios, such as severe halos from direct headlights, which can corrupt the attenuation signal. Future work will focus on explicitly modeling these complex lighting phenomena. Ultimately, PhysDepth validates a new, physically-grounded paradigm for MDE, paving the way for next-generation models that can operate reliably in the challenging conditions of the real world.
{
    \small
    \bibliographystyle{ieeenat_fullname}
    \bibliography{main}

@String(CVPR= {IEEE Conf. Comput. Vis. Pattern Recog.})

@String(ICCV= {Int. Conf. Comput. Vis.})

@String(ECCV= {Eur. Conf. Comput. Vis.})

@String(ICASSP=	{ICASSP})

@String(CVPR  = {CVPR})

@String(ICCV  = {ICCV})

@String(ECCV  = {ECCV})

@inproceedings{wang2021regularizing,
  title={Regularizing nighttime weirdness: Efficient self-supervised monocular depth estimation in the dark},
  author={Wang, Kun and Zhang, Zhenyu and Yan, Zhiqiang and Li, Xiang and Xu, Baobei and Li, Jun and Yang, Jian},
  booktitle={Proceedings of the IEEE/CVF international conference on computer vision},
  pages={16055--16064},
  year={2021}
}

@article{ranftl2020towards,
  title={Towards robust monocular depth estimation: Mixing datasets for zero-shot cross-dataset transfer},
  author={Ranftl, Ren{\'e} and Lasinger, Katrin and Hafner, David and Schindler, Konrad and Koltun, Vladlen},
  journal={IEEE transactions on pattern analysis and machine intelligence},
  volume={44},
  number={3},
  pages={1623--1637},
  year={2020},
  publisher={IEEE}
}

@inproceedings{gasperini2021r4dyn,
  title={R4dyn: Exploring radar for self-supervised monocular depth estimation of dynamic scenes},
  author={Gasperini, Stefano and Koch, Patrick and Dallabetta, Vinzenz and Navab, Nassir and Busam, Benjamin and Tombari, Federico},
  booktitle={2021 International Conference on 3D Vision (3DV)},
  pages={751--760},
  year={2021},
  organization={IEEE}
}

@inproceedings{thanh2024depthsegnet24,
  title={DepthSegNet24: A Label-Free Model for Robust Day-Night Depth and Semantics},
  author={Thanh, Phan Thi Huyen and Nguyen, Minh Huy Vu and Tran, Trung Thai and Pham, Tran Vu and Nguyen, Duc Dung and Duy, Truong Vinh Truong and Naotake, Natori and others},
  booktitle={Proceedings of the Asian Conference on Computer Vision},
  pages={2716--2733},
  year={2024}
}

@article{cong2024srnsd,
  title={SRNSD: Structure-Regularized Night-Time Self-Supervised Monocular Depth Estimation for Outdoor Scenes},
  author={Cong, Runmin and Wu, Chunlei and Song, Xibin and Zhang, Wei and Kwong, Sam and Li, Hongdong and Ji, Pan},
  journal={IEEE Transactions on Image Processing},
  year={2024},
  publisher={IEEE}
}

@inproceedings{vankadari2023sun,
  title={When the sun goes down: Repairing photometric losses for all-day depth estimation},
  author={Vankadari, Madhu and Golodetz, Stuart and Garg, Sourav and Shin, Sangyun and Markham, Andrew and Trigoni, Niki},
  booktitle={Conference on Robot Learning},
  pages={1992--2003},
  year={2023},
  organization={PMLR}
}

@article{ba2016layer,
  title={Layer normalization},
  author={Ba, Jimmy Lei and Kiros, Jamie Ryan and Hinton, Geoffrey E},
  journal={arXiv preprint arXiv:1607.06450},
  year={2016}
}

@inproceedings{piccinelli2024unidepth,
  title={UniDepth: Universal monocular metric depth estimation},
  author={Piccinelli, Luigi and Yang, Yung-Hsu and Sakaridis, Christos and Segu, Mattia and Li, Siyuan and Van Gool, Luc and Yu, Fisher},
  booktitle={Proceedings of the IEEE/CVF Conference on Computer Vision and Pattern Recognition},
  pages={10106--10116},
  year={2024}
}

@inproceedings{liu2022convnet,
  title={A convnet for the 2020s},
  author={Liu, Zhuang and Mao, Hanzi and Wu, Chao-Yuan and Feichtenhofer, Christoph and Darrell, Trevor and Xie, Saining},
  booktitle={Proceedings of the IEEE/CVF conference on computer vision and pattern recognition},
  pages={11976--11986},
  year={2022}
}

@article{hong2024you,
  title={You Only Look Around: Learning Illumination Invariant Feature for Low-light Object Detection},
  author={Hong, Mingbo and Cheng, Shen and Huang, Haibin and Fan, Haoqiang and Liu, Shuaicheng},
  journal={Advances in Neural Information Processing Systems},
  year={2024}
}

@inproceedings{moon2024ground,
  title={From-Ground-To-Objects: Coarse-to-Fine Self-supervised Monocular Depth Estimation of Dynamic Objects with Ground Contact Prior},
  author={Moon, Jaeho and Bello, Juan Luis Gonzalez and Kwon, Byeongjun and Kim, Munchurl},
  booktitle={Proceedings of the IEEE/CVF Conference on Computer Vision and Pattern Recognition},
  pages={10519--10529},
  year={2024}
}

@inproceedings{li2022physically,
  title={Physically-based editing of indoor scene lighting from a single image},
  author={Li, Zhengqin and Shi, Jia and Bi, Sai and Zhu, Rui and Sunkavalli, Kalyan and Ha{\v{s}}an, Milo{\v{s}} and Xu, Zexiang and Ramamoorthi, Ravi and Chandraker, Manmohan},
  booktitle={European Conference on Computer Vision},
  pages={555--572},
  year={2022},
  organization={Springer}
}

@inproceedings{wang2021learning,
  title={Learning indoor inverse rendering with 3d spatially-varying lighting},
  author={Wang, Zian and Philion, Jonah and Fidler, Sanja and Kautz, Jan},
  booktitle={Proceedings of the IEEE/CVF International Conference on Computer Vision},
  pages={12538--12547},
  year={2021}
}

@article{wang2009all,
author = {Wang, Jiaping and Ren, Peiran and Gong, Minmin and Snyder, John and Guo, Baining},
title = {All-frequency rendering of dynamic, spatially-varying reflectance},
year = {2009},
issue_date = {December 2009},
publisher = {Association for Computing Machinery},
address = {New York, NY, USA},
volume = {28},
number = {5},
issn = {0730-0301},
url = {https://doi.org/10.1145/1618452.1618479},
doi = {10.1145/1618452.1618479},
journal = {ACM Trans. Graph.},
month = dec,
pages = {1–10},
numpages = {10}
}

@misc{attenuation,
  author = {https://wiki.ogre3d.org/},
  year = {2011},
  url = {https://wiki.ogre3d.org/tiki-index.php?page=-Point+Light+Attenuation},
  urldate = {February 22, 2011},
  title = {Point Light Attenuation}
}

@INPROCEEDINGS{zheng2023steps,
  author={Zheng, Yupeng and Zhong, Chengliang and Li, Pengfei and Gao, Huan-ang and Zheng, Yuhang and Jin, Bu and Wang, Ling and Zhao, Hao and Zhou, Guyue and Zhang, Qichao and Zhao, Dongbin},
  booktitle={2023 IEEE International Conference on Robotics and Automation (ICRA)}, 
  title={STEPS: Joint Self-supervised Nighttime Image Enhancement and Depth Estimation}, 
  year={2023},
  volume={},
  number={},
  pages={4916-4923},
  doi={10.1109/ICRA48891.2023.10160708}}

@inproceedings{ummenhofer2017demon,
  title={Demon: Depth and motion network for learning monocular stereo},
  author={Ummenhofer, Benjamin and Zhou, Huizhong and Uhrig, Jonas and Mayer, Nikolaus and Ilg, Eddy and Dosovitskiy, Alexey and Brox, Thomas},
  booktitle={Proceedings of the IEEE conference on computer vision and pattern recognition},
  pages={5038--5047},
  year={2017}
}

@article{RobotCarDatasetIJRR,
author = {Maddern, Will and Pascoe, Geoffrey and Linegar, Chris and Newman, Paul},
title = {1 year, 1000 km: The Oxford RobotCar dataset},
year = {2017},
issue_date = {Jan 2017},
publisher = {Sage Publications, Inc.},
address = {USA},
volume = {36},
number = {1},
issn = {0278-3649},
url = {https://doi.org/10.1177/0278364916679498},
doi = {10.1177/0278364916679498},
journal = {Int. J. Rob. Res.},
month = jan,
pages = {3–15},
numpages = {13}
}

@inproceedings{caesar2020nuscenes,
  title={nuscenes: A multimodal dataset for autonomous driving},
  author={Caesar, Holger and Bankiti, Varun and Lang, Alex H and Vora, Sourabh and Liong, Venice Erin and Xu, Qiang and Krishnan, Anush and Pan, Yu and Baldan, Giancarlo and Beijbom, Oscar},
  booktitle={Proceedings of the IEEE/CVF conference on computer vision and pattern recognition},
  pages={11621--11631},
  year={2020}
}

@inproceedings{zhou2017unsupervised,
  title={Unsupervised learning of depth and ego-motion from video},
  author={Zhou, Tinghui and Brown, Matthew and Snavely, Noah and Lowe, David G},
  booktitle={Proceedings of the IEEE conference on computer vision and pattern recognition},
  pages={1851--1858},
  year={2017}
}

@inproceedings{godard2017unsupervised,
  title={Unsupervised monocular depth estimation with left-right consistency},
  author={Godard, Cl{\'e}ment and Mac Aodha, Oisin and Brostow, Gabriel J},
  booktitle={Proceedings of the IEEE conference on computer vision and pattern recognition},
  pages={270--279},
  year={2017}
}

@inproceedings{vankadari2020unsupervised,
  title={Unsupervised monocular depth estimation for night-time images using adversarial domain feature adaptation},
  author={Vankadari, Madhu and Garg, Sourav and Majumder, Anima and Kumar, Swagat and Behera, Ardhendu},
  booktitle={European Conference on Computer Vision},
  pages={443--459},
  year={2020},
  organization={Springer}
}

@inproceedings{sharma2020nighttime,
  title={Nighttime stereo depth estimation using joint translation-stereo learning: Light effects and uninformative regions},
  author={Sharma, Aashish and Cheong, Loong-Fah and Heng, Lionel and Tan, Robby T},
  booktitle={2020 International Conference on 3D Vision (3DV)},
  pages={23--31},
  year={2020},
  organization={IEEE}
}

@inproceedings{geiger2012we,
  title={Are we ready for autonomous driving? the kitti vision benchmark suite},
  author={Geiger, Andreas and Lenz, Philip and Urtasun, Raquel},
  booktitle={2012 IEEE Conference on Computer Vision and Pattern Recognition},
  pages={3354--3361},
  year={2012},
  organization={IEEE}
}

@inproceedings{paszke2019pytorch,
author = {Paszke, Adam and Gross, Sam and Massa, Francisco and Lerer, Adam and Bradbury, James and Chanan, Gregory and Killeen, Trevor and Lin, Zeming and Gimelshein, Natalia and Antiga, Luca and Desmaison, Alban and K\"{o}pf, Andreas and Yang, Edward and DeVito, Zach and Raison, Martin and Tejani, Alykhan and Chilamkurthy, Sasank and Steiner, Benoit and Fang, Lu and Bai, Junjie and Chintala, Soumith},
title = {PyTorch: an imperative style, high-performance deep learning library},
year = {2019},
publisher = {Curran Associates Inc.},
address = {Red Hook, NY, USA},
booktitle = {Proceedings of the 33rd International Conference on Neural Information Processing Systems},
articleno = {721},
numpages = {12}
}

@article{kingma2014adam,
  title={Adam: A method for stochastic optimization},
  author={Kingma, Diederik P and Ba, Jimmy},
  journal={arXiv preprint arXiv:1412.6980},
  year={2014}
}

@article{eigen2014depth,
  title={Depth map prediction from a single image using a multi-scale deep network},
  author={Eigen, David and Puhrsch, Christian and Fergus, Rob},
  journal={Advances in neural information processing systems},
  volume={27},
  year={2014}
}

@inproceedings{lu2021alternative,
  title={An alternative of lidar in nighttime: Unsupervised depth estimation based on single thermal image},
  author={Lu, Yawen and Lu, Guoyu},
  booktitle={Proceedings of the IEEE/CVF Winter Conference on Applications of Computer Vision},
  pages={3833--3843},
  year={2021}
}

@inproceedings{bhat2021adabins,
  title={Adabins: Depth estimation using adaptive bins},
  author={Bhat, Shariq Farooq and Alhashim, Ibraheem and Wonka, Peter},
  booktitle={Proceedings of the IEEE/CVF Conference on Computer Vision and Pattern Recognition},
  pages={4009--4018},
  year={2021}
}

@inproceedings{spencer2020defeat,
  title={Defeat-net: General monocular depth via simultaneous unsupervised representation learning},
  author={Spencer, Jaime and Bowden, Richard and Hadfield, Simon},
  booktitle={Proceedings of the IEEE/CVF Conference on Computer Vision and Pattern Recognition},
  pages={14402--14413},
  year={2020}
}

@inproceedings{shu2020feature,
  title={Feature-metric loss for self-supervised learning of depth and egomotion},
  author={Shu, Chang and Yu, Kun and Duan, Zhixiang and Yang, Kuiyuan},
  booktitle={European Conference on Computer Vision},
  pages={572--588},
  year={2020},
  organization={Springer}
}

@inproceedings{peng2021excavating,
  title={Excavating the potential capacity of self-supervised monocular depth estimation},
  author={Peng, Rui and Wang, Ronggang and Lai, Yawen and Tang, Luyang and Cai, Yangang},
  booktitle={Proceedings of the IEEE/CVF International Conference on Computer Vision},
  pages={15560--15569},
  year={2021}
}

@article{park2021diffnet,
  title={DIFFnet: Diffusion Parameter Mapping Network Generalized for Input Diffusion Gradient Schemes and b-Value},
  author={Park, Juhyung and Jung, Woojin and Choi, Eun-Jung and Oh, Se-Hong and Jang, Jinhee and Shin, Dongmyung and An, Hongjun and Lee, Jongho},
  journal={IEEE Transactions on Medical Imaging},
  volume={41},
  number={2},
  pages={491--499},
  year={2021},
  publisher={IEEE}
}

@InProceedings{feng2022disentangling,
author="Feng, Ziyue
and Yang, Liang
and Jing, Longlong
and Wang, Haiyan
and Tian, YingLi
and Li, Bing",
editor="Avidan, Shai
and Brostow, Gabriel
and Ciss{\'e}, Moustapha
and Farinella, Giovanni Maria
and Hassner, Tal",
title="Disentangling Object Motion and Occlusion for Unsupervised Multi-frame Monocular Depth",
booktitle="Computer Vision -- ECCV 2022",
year="2022",
publisher="Springer Nature Switzerland",
address="Cham",
pages="228--244",
isbn="978-3-031-19824-3"
}

@inproceedings{yan2021channel,
  title={Channel-Wise Attention-Based Network for Self-Supervised Monocular Depth Estimation},
  author={Yan, Jiaxing and Zhao, Hong and Bu, Penghui and Jin, YuSheng},
  booktitle={2021 International Conference on 3D Vision (3DV)},
  pages={464--473},
  year={2021},
  organization={IEEE}
}

@article{li2022depthformer,
  title={Depthformer: Exploiting long-range correlation and local information for accurate monocular depth estimation},
  author={Li, Zhenyu and Chen, Zehui and Liu, Xianming and Jiang, Junjun},
  journal={Machine Intelligence Research},
  volume={20},
  number={6},
  pages={837--854},
  year={2023},
  publisher={Springer}
}

@inproceedings{gasperini2023robust,
  title={Robust monocular depth estimation under challenging conditions},
  author={Gasperini, Stefano and Morbitzer, Nils and Jung, HyunJun and Navab, Nassir and Tombari, Federico},
  booktitle={Proceedings of the IEEE/CVF International Conference on Computer Vision},
  pages={8177--8186},
  year={2023}
}

@inproceedings{guizilini20203d,
  title={3d packing for self-supervised monocular depth estimation},
  author={Guizilini, Vitor and Ambrus, Rares and Pillai, Sudeep and Raventos, Allan and Gaidon, Adrien},
  booktitle={Proceedings of the IEEE/CVF conference on computer vision and pattern recognition},
  pages={2485--2494},
  year={2020}
}

@inproceedings{wang2023planedepth,
  title={PlaneDepth: Self-Supervised Depth Estimation via Orthogonal Planes},
  author={Wang, Ruoyu and Yu, Zehao and Gao, Shenghua},
  booktitle={Proceedings of the IEEE/CVF Conference on Computer Vision and Pattern Recognition},
  pages={21425--21434},
  year={2023}
}

@ARTICLE{10696933,
  author={Cong, Runmin and Wu, Chunlei and Song, Xibin and Zhang, Wei and Kwong, Sam and Li, Hongdong and Ji, Pan},
  journal={IEEE Transactions on Image Processing}, 
  title={SRNSD: Structure-Regularized Night-Time Self-Supervised Monocular Depth Estimation for Outdoor Scenes}, 
  year={2024},
  volume={33},
  number={},
  pages={5538-5550},
  doi={10.1109/TIP.2024.3465034}}

@ARTICLE{10539960,
  author={Hou, Shengyu and Fu, Mengyin and Wang, Rongchuan and Yang, Yi and Song, Wenjie},
  journal={IEEE Transactions on Circuits and Systems for Video Technology}, 
  title={Self-Supervised Monocular Depth Estimation for All-Day Images Based on Dual-Axis Transformer}, 
  year={2024},
  volume={34},
  number={10},
  pages={9939-9953},
  doi={10.1109/TCSVT.2024.3406043}}

@article{dong2022towards,
  title={Towards real-time monocular depth estimation for robotics: A survey},
  author={Dong, Xingshuai and Garratt, Matthew A and Anavatti, Sreenatha G and Abbass, Hussein A},
  journal={IEEE Transactions on Intelligent Transportation Systems},
  volume={23},
  number={10},
  pages={16940--16961},
  year={2022},
  publisher={IEEE}
}

@article{zhou2019does,
  title={Does computer vision matter for action?},
  author={Zhou, Brady and Kr{\"a}henb{\"u}hl, Philipp and Koltun, Vladlen},
  journal={Science Robotics},
  volume={4},
  number={30},
  pages={eaaw6661},
  year={2019},
  publisher={American Association for the Advancement of Science}
}

@inproceedings{park2021pseudo,
  title={Is pseudo-lidar needed for monocular 3d object detection?},
  author={Park, Dennis and Ambrus, Rares and Guizilini, Vitor and Li, Jie and Gaidon, Adrien},
  booktitle={Proceedings of the IEEE/CVF International Conference on Computer Vision},
  pages={3142--3152},
  year={2021}
}

@inproceedings{wang2019pseudo,
  title={Pseudo-lidar from visual depth estimation: Bridging the gap in 3d object detection for autonomous driving},
  author={Wang, Yan and Chao, Wei-Lun and Garg, Divyansh and Hariharan, Bharath and Campbell, Mark and Weinberger, Kilian Q},
  booktitle={Proceedings of the IEEE/CVF conference on computer vision and pattern recognition},
  pages={8445--8453},
  year={2019}
}

@inproceedings{vankadari2024dusk,
  title={Dusk till dawn: Self-supervised nighttime stereo depth estimation using visual foundation models},
  author={Vankadari, Madhu and Hodgson, Samuel and Shin, Sangyun and Zhou, Kaichen and Markham, Andrew and Trigoni, Niki},
  booktitle={2024 IEEE International Conference on Robotics and Automation (ICRA)},
  pages={17976--17982},
  year={2024},
  organization={IEEE}
}

@article{rajapaksha2024deep,
  title={Deep learning-based depth estimation methods from monocular image and videos: A comprehensive survey},
  author={Rajapaksha, Uchitha and Sohel, Ferdous and Laga, Hamid and Diepeveen, Dean and Bennamoun, Mohammed},
  journal={ACM Computing Surveys},
  volume={56},
  number={12},
  pages={1--51},
  year={2024},
  publisher={ACM New York, NY}
}

@inproceedings{liu2021self,
  title={Self-supervised monocular depth estimation for all day images using domain separation},
  author={Liu, Lina and Song, Xibin and Wang, Mengmeng and Liu, Yong and Zhang, Liangjun},
  booktitle={Proceedings of the IEEE/CVF international conference on computer vision},
  pages={12737--12746},
  year={2021}
}

@article{zhao2022unsupervised,
  title={Unsupervised monocular depth estimation in highly complex environments},
  author={Zhao, Chaoqiang and Tang, Yang and Sun, Qiyu},
  journal={IEEE Transactions on Emerging Topics in Computational Intelligence},
  volume={6},
  number={5},
  pages={1237--1246},
  year={2022},
  publisher={IEEE}
}

@inproceedings{saunders2023self,
  title={Self-supervised monocular depth estimation: Let's talk about the weather},
  author={Saunders, Kieran and Vogiatzis, George and Manso, Luis J},
  booktitle={Proceedings of the IEEE/CVF International Conference on Computer Vision},
  pages={8907--8917},
  year={2023}
}

@article{he2010single,
  title={Single image haze removal using dark channel prior},
  author={He, Kaiming and Sun, Jian and Tang, Xiaoou},
  journal={IEEE transactions on pattern analysis and machine intelligence},
  volume={33},
  number={12},
  pages={2341--2353},
  year={2010},
  publisher={IEEE}
}

@inproceedings{tan2008visibility,
  title={Visibility in bad weather from a single image},
  author={Tan, Robby T},
  booktitle={2008 IEEE conference on computer vision and pattern recognition},
  pages={1--8},
  year={2008},
  organization={IEEE}
}

@inproceedings{10.1145/1399504.1360671,
author = {Fattal, Raanan},
title = {Single image dehazing},
year = {2008},
isbn = {9781450301121},
publisher = {Association for Computing Machinery},
address = {New York, NY, USA},
url = {https://doi.org/10.1145/1399504.1360671},
doi = {10.1145/1399504.1360671},
booktitle = {ACM SIGGRAPH 2008 Papers},
articleno = {72},
numpages = {9},
location = {Los Angeles, California},
series = {SIGGRAPH '08}
}

@inproceedings{narasimhan2000chromatic,
  title={Chromatic framework for vision in bad weather},
  author={Narasimhan, Srinivasa G and Nayar, Shree K},
  booktitle={Proceedings IEEE Conference on Computer Vision and Pattern Recognition. CVPR 2000 (Cat. No. PR00662)},
  volume={1},
  pages={598--605},
  year={2000},
  organization={IEEE}
}

@article{narasimhan2002vision,
  title={Vision and the atmosphere},
  author={Narasimhan, Srinivasa G and Nayar, Shree K},
  journal={International journal of computer vision},
  volume={48},
  number={3},
  pages={233--254},
  year={2002},
  publisher={Springer}
}

@inproceedings{godard2019digging,
  title={Digging into self-supervised monocular depth estimation},
  author={Godard, Cl{\'e}ment and Mac Aodha, Oisin and Firman, Michael and Brostow, Gabriel J},
  booktitle={Proceedings of the IEEE/CVF international conference on computer vision},
  pages={3828--3838},
  year={2019}
}

@inproceedings{zhang2025lightweight,
  title={Lightweight Self-Supervised Monocular Depth Estimation for All-Day Scenes Using Generative Adversarial Network},
  author={Zhang, Junding and Rao, Di and Akoudad, Youssef and Gao, Wei and Chen, Jie},
  booktitle={ICASSP 2025-2025 IEEE International Conference on Acoustics, Speech and Signal Processing (ICASSP)},
  pages={1--5},
  year={2025},
  organization={IEEE}
}

@InProceedings{Yan_2025_CVPR,
    author    = {Yan, Weilong and Li, Ming and Li, Haipeng and Shao, Shuwei and Tan, Robby T.},
    title     = {Synthetic-to-Real Self-supervised Robust Depth Estimation via Learning with Motion and Structure Priors},
    booktitle = {Proceedings of the IEEE/CVF Conference on Computer Vision and Pattern Recognition (CVPR)},
    month     = {June},
    year      = {2025},
    pages     = {21880-21890}
}

@InProceedings{Zhang_2025_ICCV,
    author    = {Zhang, Wenyao and Liu, Hongsi and Li, Bohan and He, Jiawei and Qi, Zekun and Wang, Yunnan and Zhao, Shengyang and Yu, Xinqiang and Zeng, Wenjun and Jin, Xin},
    title     = {Hybrid-grained Feature Aggregation with Coarse-to-fine Language Guidance for Self-supervised Monocular Depth Estimation},
    booktitle = {Proceedings of the IEEE/CVF International Conference on Computer Vision (ICCV)},
    month     = {October},
    year      = {2025},
    pages     = {6678-6692}
}

@InProceedings{Chung_2025_ICCV,
    author    = {Chung, Younjoon and Park, Hyoungseob and Rim, Patrick and Zhang, Xiaoran and He, Jihe and Zeng, Ziyao and Cicek, Safa and Hong, Byung-Woo and Duncan, James S. and Wong, Alex},
    title     = {ETA: Energy-based Test-time Adaptation for Depth Completion},
    booktitle = {Proceedings of the IEEE/CVF International Conference on Computer Vision (ICCV)},
    month     = {October},
    year      = {2025},
    pages     = {6001-6012}
}

@inproceedings{zeng2025depthdark,
  title={DepthDark: Robust Monocular Depth Estimation for Low-Light Environments},
  author={Zeng, Longjian and Zhu, Zunjie and Lu, Rongfeng and Lu, Ming and Zheng, Bolun and Yan, Chenggang and Xue, Anke},
  booktitle={Proceedings of the 33rd ACM International Conference on Multimedia},
  pages={11239--11248},
  year={2025}
}

@inproceedings{yan2025synthetic,
  title={Synthetic-to-real self-supervised robust depth estimation via learning with motion and structure priors},
  author={Yan, Weilong and Li, Ming and Li, Haipeng and Shao, Shuwei and Tan, Robby T},
  booktitle={Proceedings of the Computer Vision and Pattern Recognition Conference},
  pages={21880--21890},
  year={2025}
}
}
\clearpage
\setcounter{page}{1}
\maketitlesupplementary

\begin{algorithm*}[!ht]
\caption{Covariance Computation Computation}
\label{alg:blds}
\begin{algorithmic}[1]
\Require MDE model $M$
\Require Test dataset $\mathcal{D} = \{(I^{(j)}, D_{gt}^{(j)})\}_{j=1}^{K}$
\Require Perturbation vector $\mathbf{B} = [\beta_0, \dots, \beta_N]$ (extinction coefficients)
\Require Nocturnal Airlight $A$ (e.g., $[0.1, 0.1, 0.1]$), Error function $\text{Error}(\cdot, \cdot)$ (AbsRel)
\Require Average Covariance score $\overline{\text{Cov}}$

\State Initialize $\text{score\_list} \gets []$
\For{each $(I^{(j)}, D_{gt}^{(j)}) \in \mathcal{D}$} \Comment{Loop over all K images in test set}
    \State Initialize error vector $\mathbf{E}^{(j)} \gets []$
    \For{each $\beta_i \in \mathbf{B}$} \Comment{Loop over N perturbation levels}
        \Comment{\textbf{Step 1:} Generate Perturbed Image}
        \State $t_i \gets \exp(-\beta_i \cdot D_{gt}^{(j)})$ \Comment{Calculate transmission map}
        \State $I'_i \gets I^{(j)} \cdot t_i + A \cdot (1 - t_i)$ \Comment{Synthesize attenuated image}
        
        \Comment{\textbf{Step 2:} Model Inference}
        \State $D_{pred, i} \gets M(I'_i)$ \Comment{Get model's prediction}
        
        \Comment{\textbf{Step 3:} Calculate Error}
        \State $\mathcal{E}_i \gets \text{Error}(D_{pred, i}, D_{gt}^{(j)})$
        \State Append $\mathcal{E}_i$ to $\mathbf{E}^{(j)}$
    \EndFor
    
    \Comment{\textbf{Step 4:} Compute Score for image $j$}
    \State $\rho_j \gets \text{PearsonCorrelation}(\mathbf{B}, \mathbf{E}^{(j)})$ \Comment{$\rho(\mathbf{B}, \mathbf{E}^{(j)})$}
    \State Append $\rho_j$ to $\text{score\_list}$
\EndFor

\State $\overline{\text{Cov}} \gets \text{Mean}(\text{score\_list})$ \Comment{Average score over all images}
\State \Return $\overline{\text{Cov}}$
\end{algorithmic}
\end{algorithm*}

\section{Supplementary Material Introduction}\label{sec:suppl intro}

This appendix includes additional details and results to support our manuscript. In particular, this appendix is organized as follows:
\begin{itemize}

\item In Section \ref{sec:BLDS Alg}, we provide the formal pseudocode detailing the step-by-step computation of covariance $Cov$ between a model's error metric and atmosphere attenuation $\beta$

\item In Section \ref{sec:Encoder Design}, we discuss more details about the PhysDepth

\item In Section \ref{sec:suppl DataSet Details}, we discuss more details about the datasets.

\item In Section \ref{sec:Additional Quantitative Results}, we show more quantitative results.

\item In Section \ref{sec:Additional Qualitative Results}, we present additional qualitative results.

\item In Section \ref{sec:daytime}, we verify the performance of PhysDepth on daytime dataset.

\item In Section \ref{sec:Failure Cases}, we review failure cases.
\end{itemize}

 \section{Covariance Computation}\label{sec:BLDS Alg}
 This section presents Algorithm \ref{alg:blds}, which provides the pseudocode detailing the step-by-step computation of the Covariance used in our empirical study.

\textbf{Rationale and Hypothesis.}
Our core hypothesis is that a model's response to controlled attenuation reveals its underlying strategy:

1) Heuristic-Follower (Non-Physical): A model that relies on simple heuristics (e.g., "brighter light = closer") or texture cues will perceive attenuation as noise. As the attenuation coefficient increases, its primary cues are destroyed, and its prediction error will increase.

2) Physics-Informed (Physical): A model that has learned to invert the Beer-Lambert law will treat attenuation as a signal. As the attenuation coefficient increases, it gains a stronger, valid cue for depth (i.e., it correctly identifies that dimmer objects, relative to their known class, are farther away), and its prediction error will remain stable or decrease.

\textbf{Ablation for Covariance:} A potential flaw of the Covariance is that it may unfairly favor our model, which is trained on a loss derived from the same physical law. 
To disentangle the contribution of our RCA loss from the simpler effect of merely exposing the model to attenuated images, we conduct a crucial ablation study.
We compare three models:

\begin{itemize}
    \item Model A (Baseline): The standard  ViT-B/32 model trained with only the photometric loss.
    \item Model B (Augment-Only): The baseline  ViT-B/32 model trained with the \cref{eq:blds} applied as a random data augmentation.
    \item Model C (Ours): Our full PhysDepth model ( ViT-B/32+ PPM + RCA), trained with our proposed RCA loss but without covariance augmentation.
\end{itemize}

We then set $\beta = 0.025$ and evaluate all three models on the Covariance, with results in Table \ref{tab:blds_ablation}. 
It clearly demonstrates the limitation of non-physical approaches. The Baseline (Model A) is brittle, with a positive Covariance of +0.180. Applying data augmentation (Model B) provides some robustness, lowering the score to +0.142, but the model's strategy remains "brittle" ($Cov > 0$), as it still treats attenuation as noise. In sharp contrast, our full model (Model C) achieves a score of $\mathbf{-0.040}$. This result is critical for two reasons: 1.  The score is negative ($Cov < 0$). It demonstrates that our model has successfully learned to invert the Beer-Lambert law and use attenuation as a valid (albeit subtle) signal for depth, aligning with the "Physics-Informed" strategy.  The score is near-zero ($Cov \approx 0.0$), which proves our model has also learned a near-perfectly robust and decoupled strategy, performing invariantly to attenuation.
This is the paper's key finding: simple data augmentation (Model B) can only learn some invariance, and its strategy remains brittle (positive score). Our RCA loss forces the model to learn the true underlying physical principle, proven by its ability to achieve a robust, negative-scoring strategy.

\begin{table}[ht]
\centering
\caption{
Ablation study isolating the contribution of the RCA loss. We compare  Covariance across three models: (A) the baseline, (B) the baseline trained with Covariance perturbations as data augmentation, and (C) our full model with the RCA loss. The negative score for our model demonstrates it has learned to invert the attenuation, while others still treat it as noise.
}
\label{tab:blds_ablation}
\resizebox{\columnwidth}{!}{
\begin{tabular}{l c c c}
\toprule
\textbf{Method} & \textbf{RCA Loss} & \textbf{Cov Augment.} & \textbf{Cov (AbsRel)} \\
\midrule
Model A (Baseline)      & $\times$ & $\times$ & $+0.180$ \\ 
Model B (Augment-Only)  & $\times$ & \checkmark & $+0.142$ \\ 
\textbf{Model C (Ours)} & \checkmark & $\times$ & $-0.040$ \\ 
\bottomrule
\end{tabular}
} 
\end{table}

\section{PhysDepth Design}\label{sec:Encoder Design}

In our base MDE model, the encoder is followed by a self-attention and a series of convolutional upsampling layers to form the decoder
Regarding the prediction layers, like Zhou et al. \cite{zhou2017unsupervised}, we use
$ \frac{1}{\alpha \cdot \text{Sigmoid}(x) + \beta} $ with $ \alpha = 10 $ and $ \beta = 0.0125 $ to constrain the predicted depth to be always positive within a reasonable range (e.g., $ 80 $ meters). 
In addition, we use a Self-Attention  module between the encoder and decoder to further improve the accuracy of the baseline.

\textbf{Encoder:} Following follow Piccinelli et al. \cite{piccinelli2024unidepth}, for ViT and ConvNeXt we remove all classification-specific layers (e.g., global pooling, fully connected, and softmax) from each. Then we first take the encoder output, flatten it, and apply LayerNorm \cite{ba2016layer}. We then project the feature to a 1024-dimensional channel space  using a linear layer and bilinearly interpolate the result to a same resolution of H and W as ViT encoder's output dimension

\textbf{Pose Estimation CNN:}
The pose estimation network learns a pose transformation between two consecutive frames, which serves as a part of our self-supervised signals for depth estimation. 
We follow the same network in SfMLearner \cite{zhou2017unsupervised}, in which we concatenate two consecutive frames along the channel direction. 
The concatenated result is fed into the Pose Estimation CNN.
The outputs are the relative poses between each of the source views and the target view.
Specifically, the network consists of 2 convolutions followed by a $ 1 \times 1 $ convolution with $ 6 \times 1 $ output channels. One channel corresponds to 3 Euler angles and a 3-D translation for each source view. All convolution layers are followed by ReLU except for the last layer, where no nonlinear activation is applied.
Since we use photometric reconstruction loss function $ \mathcal{L}_\text{p} $ from SfMLearner \cite{zhou2017unsupervised} during the training processes, which implies that the rigid assumptions in SfMLearner \cite{zhou2017unsupervised} also apply to our model.

\textbf{Weak Velocity Supervision:} We follow \cite{guizilini20203d,gasperini2023robust} to use a weak velocity supervision $\mathcal{L}_v$ for scale awareness and ensure consistent predictions when the photometric
consistency does not hold. 
$\mathcal{L}_v$ is defined in Equation \ref{eq:lv}

\begin{equation}
    \label{eq:lv}
    \mathcal{L}_v (\hat{T}_{t \rightarrow s},{T}_{t \rightarrow s}) = \big| ||{\hat{T}_{t \rightarrow s}}||_2- ||{{T}_{t \rightarrow s}}||_2 \big|
\end{equation}

$\hat{T}_{t \rightarrow s}$ and ${T}_{t \rightarrow s}$ are the predicted and ground truth
pose translations, respectively.
We can easily obtain them from the available odometry information, through the ego vehicle speed and the time interval across frames.

\section{DataSet Details}\label{sec:suppl DataSet Details}

\textbf{Oxford RobotCar-Night:}
The RobotCar-Night dataset contains images from the 2014-12-16-18-44-24 folder, which includes 6000 training images and 411 test images.
All images are cropped to $1152 \times 672$.

\textbf{nuScenes-Night and nuScenes-Rain:} From the original nuScenes dataset \cite{caesar2020nuscenes}, Wang et al. \cite{wang2021regularizing} created the nuScenes-Night dataset split, which contains over 10000 training images and 500 test samples. All images are cropped to $1536 \times 768$.
All images are in $1242 \times 375$.

\begin{table}[ht]
    \centering
    \footnotesize
    \caption{\textbf{Ablation Study:} We test PhysDepth and Plugin-enhanced Models on all images from nuScenes-Night dataset \cite{caesar2020nuscenes} in 40 meters depth range. The best results are in \textbf{bold}.}    \vspace{-2mm}
    \resizebox{\columnwidth}{!}{
    \begin{tabular}{l|l|cc|c}
    \toprule
    \multicolumn{1}{c|}{} & \multicolumn{1}{c|}{Methods} & \multicolumn{1}{c|}{Abs Rel} & \multicolumn{1}{c|}{RMSE} & \multicolumn{1}{c}{$\delta<1.25$} \\
    \multicolumn{1}{c|}{} & \textbf{} & \multicolumn{2}{c|}{lower's better} & \multicolumn{1}{c}{higher's better} \\
    \hline
    
    \multirow{3}{*}{SOTA} & Monodepth2 & \multicolumn{1}{c|}{0.116} & \multicolumn{1}{c|}{3.701} & \multicolumn{1}{c}{0.876} \\ \cline{2-5}
    
     & md4All-AD & \multicolumn{1}{c|}{0.121} & \multicolumn{1}{c|}{3.851} & \multicolumn{1}{c}{0.870} \\ \cline{2-5}
    
     & md4All-DD & \multicolumn{1}{c|}{0.109} & \multicolumn{1}{c|}{3.606} & \multicolumn{1}{c}{0.883} \\ \hline
    
    \multirow{4}{*}{\makecell[l]{Plugin\\Enhanced\\Models}} & PhysDepth+DepthDark & \multicolumn{1}{c|}{0.111} & \multicolumn{1}{c|}{3.595} & \multicolumn{1}{c}{0.905} \\ \cline{2-5}
    
     & PhysDepth+ConvNeXt & \multicolumn{1}{c|}{0.107} & \multicolumn{1}{c|}{3.593} & \multicolumn{1}{c}{0.908} \\ \cline{2-5}
    
     & PhysDepth+S2R & \multicolumn{1}{c|}{0.103} & \multicolumn{1}{c|}{3.591} & \multicolumn{1}{c}{0.910} \\ \cline{2-5}
    
     & PhysDepth+ViT-B/16 & \multicolumn{1}{c|}{0.101} & \multicolumn{1}{c|}{3.589} & \multicolumn{1}{c}{0.908} \\ \hline
    
    Ours & \textbf{PhysDepth} & \multicolumn{1}{c|}{\cellcolor{db}\textbf{0.098}} & \multicolumn{1}{c|}{\cellcolor{db}\textbf{3.586}} & \multicolumn{1}{c}{\cellcolor{db}\textbf{0.906}} \\
    
    \bottomrule
    
    \end{tabular}
    }
    \label{tab:40m nuScenes}
\end{table}

\begin{table}[ht]
    \centering
    \footnotesize
    \caption{\textbf{Ablation Study:} We test PhysDepth and Plugin-enhanced Models on all images from nuScenes-Night dataset \cite{caesar2020nuscenes} in 50 meters depth range. The best results are in \textbf{bold}.}    \vspace{-2mm}
    \resizebox{\columnwidth}{!}{
    \begin{tabular}{l|l|cc|c}
    \toprule
    \multicolumn{1}{c|}{} & \multicolumn{1}{c|}{Methods} & \multicolumn{1}{c|}{Abs Rel} & \multicolumn{1}{c|}{RMSE} & \multicolumn{1}{c}{$\delta<1.25$} \\
    \multicolumn{1}{c|}{} & \textbf{} & \multicolumn{2}{c|}{lower's better} & \multicolumn{1}{c}{higher's better} \\
    \hline
    
    \multirow{3}{*}{SOTA} & Monodepth2 & \multicolumn{1}{c|}{0.136} & \multicolumn{1}{c|}{5.645} & \multicolumn{1}{c}{0.869} \\ \cline{2-5}
    
     & md4All-AD & \multicolumn{1}{c|}{0.131} & \multicolumn{1}{c|}{5.728} & \multicolumn{1}{c}{0.865} \\ \cline{2-5}
    
     & md4All-DD & \multicolumn{1}{c|}{0.122} & \multicolumn{1}{c|}{5.413} & \multicolumn{1}{c}{0.876} \\ \hline
    
    \multirow{4}{*}{\makecell[l]{Plugin\\Enhanced\\Models}} & PhysDepth+DepthDark & \multicolumn{1}{c|}{0.115} & \multicolumn{1}{c|}{5.417} & \multicolumn{1}{c}{0.874} \\ \cline{2-5}
    
     & PhysDepth+ConvNeXt & \multicolumn{1}{c|}{0.117} & \multicolumn{1}{c|}{5.416} & \multicolumn{1}{c}{0.872} \\ \cline{2-5}
    
     & PhysDepth+S2R & \multicolumn{1}{c|}{0.116} & \multicolumn{1}{c|}{5.415} & \multicolumn{1}{c}{0.872} \\ \cline{2-5}
    
     & PhysDepth+ViT-B/16 & \multicolumn{1}{c|}{0.113} & \multicolumn{1}{c|}{5.414} & \multicolumn{1}{c}{0.876} \\ \hline
    
    Ours & \textbf{PhysDepth} & \multicolumn{1}{c|}{\cellcolor{db}\textbf{0.111}} & \multicolumn{1}{c|}{\cellcolor{db}\textbf{5.411}} & \multicolumn{1}{c}{\cellcolor{db}\textbf{0.878}} \\
    
    \bottomrule
    
    \end{tabular}
    }
    \label{tab:50m nuScenes}
\end{table}

\begin{table}[ht]
    \centering
    \footnotesize
    \caption{\textbf{Ablation Study:} We test PhysDepth and Plugin-enhanced Models on all images from nuScenes-Night dataset \cite{caesar2020nuscenes} in 60 meters depth range. The best results are in \textbf{bold}. }    \vspace{-2mm}
    \resizebox{\columnwidth}{!}{
    \begin{tabular}{l|l|cc|c}
    \toprule
    \multicolumn{1}{c|}{} & \multicolumn{1}{c|}{Methods} & \multicolumn{1}{c|}{Abs Rel} & \multicolumn{1}{c|}{RMSE} & \multicolumn{1}{c}{$\delta<1.25$} \\
    \multicolumn{1}{c|}{} & \textbf{} & \multicolumn{2}{c|}{lower's better} & \multicolumn{1}{c}{higher's better} \\
    \hline
    
    \multirow{3}{*}{SOTA} & Monodepth2 & \multicolumn{1}{c|}{0.137} & \multicolumn{1}{c|}{5.650} & \multicolumn{1}{c}{0.841} \\ \cline{2-5}
    
     & md4All-AD & \multicolumn{1}{c|}{0.135} & \multicolumn{1}{c|}{5.731} & \multicolumn{1}{c}{0.841} \\ \cline{2-5}
    
     & md4All-DD & \multicolumn{1}{c|}{0.122} & \multicolumn{1}{c|}{5.418} & \multicolumn{1}{c}{0.855} \\ \hline
    
    \multirow{4}{*}{\makecell[l]{Plugin\\Enhanced\\Models}} & PhysDepth+DepthDark & \multicolumn{1}{c|}{0.120} & \multicolumn{1}{c|}{5.424} & \multicolumn{1}{c}{0.857} \\ \cline{2-5}
    
     & PhysDepth+ConvNeXt & \multicolumn{1}{c|}{0.124} & \multicolumn{1}{c|}{5.427} & \multicolumn{1}{c}{0.854} \\ \cline{2-5}
    
     & PhysDepth+S2R & \multicolumn{1}{c|}{0.118} & \multicolumn{1}{c|}{5.420} & \multicolumn{1}{c}{0.859} \\ \cline{2-5}
    
     & PhysDepth+ViT-B/16 & \multicolumn{1}{c|}{0.117} & \multicolumn{1}{c|}{5.418} & \multicolumn{1}{c}{0.861} \\ \hline
    
    Ours & \textbf{PhysDepth} & \multicolumn{1}{c|}{\cellcolor{db}\textbf{0.115}} & \multicolumn{1}{c|}{\cellcolor{db}\textbf{5.415}} & \multicolumn{1}{c}{\cellcolor{db}\textbf{0.863}} \\
    
    \bottomrule
    
    \end{tabular}
    }
    \label{tab:60m nuScenes}
\end{table}

\section{Additional Quantitative Results}
\label{sec:Additional Quantitative Results}
\subsection{Evaluation over Different Distances}
\label{subsec:Evaluation over Different Distances using nuScenes}

\begin{table}
    \centering
    \footnotesize
    \caption{\textbf{Ablation Study:} PhysDepth and Plugin-enhanced Models on all images from the RobotCar-Night dataset \cite{wang2021regularizing} and set depth range up to 40 meters. The best results are in \textbf{bold}.}    \vspace{-2mm}
    \resizebox{\columnwidth}{!}{
    \begin{tabular}{l|l|cc|c}
    \toprule
    \multicolumn{1}{c|}{} & \multicolumn{1}{c|}{Methods} & \multicolumn{1}{c|}{Abs Rel} & \multicolumn{1}{c|}{RMSE} & \multicolumn{1}{c}{$\delta<1.25$} \\
    \multicolumn{1}{c|}{} & \textbf{} & \multicolumn{2}{c|}{lower's better} & \multicolumn{1}{c}{higher's better} \\
    \hline
    
    \multirow{3}{*}{SOTA} & Monodepth2 & \multicolumn{1}{c|}{0.302} & \multicolumn{1}{c|}{4.984} & \multicolumn{1}{c}{0.461} \\ \cline{2-5}
    
     & md4All-AD & \multicolumn{1}{c|}{0.120} & \multicolumn{1}{c|}{3.531} & \multicolumn{1}{c}{0.859} \\ \cline{2-5}
    
     & md4All-DD & \multicolumn{1}{c|}{0.121} & \multicolumn{1}{c|}{3.479} & \multicolumn{1}{c}{0.849} \\ \hline
    
    \multirow{4}{*}{\makecell[l]{Plugin\\Enhanced\\Models}} & PhysDepth+DepthDark & \multicolumn{1}{c|}{0.125} & \multicolumn{1}{c|}{3.470} & \multicolumn{1}{c}{0.847} \\ \cline{2-5}
    
     & PhysDepth+ConvNeXt & \multicolumn{1}{c|}{0.121} & \multicolumn{1}{c|}{3.467} & \multicolumn{1}{c}{0.884} \\ \cline{2-5}
    
     & PhysDepth+S2R & \multicolumn{1}{c|}{0.125} & \multicolumn{1}{c|}{3.471} & \multicolumn{1}{c}{0.844} \\ \cline{2-5}
    
     & PhysDepth+ViT-B/16 & \multicolumn{1}{c|}{0.112} & \multicolumn{1}{c|}{3.468} & \multicolumn{1}{c}{0.850} \\ \hline
    
    Ours & \textbf{PhysDepth} & \multicolumn{1}{c|}{\cellcolor{db}\textbf{0.107}} & \multicolumn{1}{c|}{\cellcolor{db}\textbf{3.445}} & \multicolumn{1}{c}{\cellcolor{db}\textbf{0.887}} \\
    
    \bottomrule
    
    \end{tabular}
    }
    \label{tab:40m robot car}
\end{table}

\begin{table}[!t]
    \centering
    \footnotesize
    \caption{\textbf{Ablation Study:} PhysDepth and Plugin-enhanced Models on all images from the RobotCar-Night dataset \cite{wang2021regularizing} and set depth range up to 60 meters. The best results are in \textbf{bold}.}    \vspace{-2mm}
    \resizebox{\columnwidth}{!}{
    \begin{tabular}{l|l|cc|c}
    \toprule
    \multicolumn{1}{c|}{} & \multicolumn{1}{c|}{Methods} & \multicolumn{1}{c|}{Abs Rel} & \multicolumn{1}{c|}{RMSE} & \multicolumn{1}{c}{$\delta<1.25$} \\
    \multicolumn{1}{c|}{} & \textbf{} & \multicolumn{2}{c|}{lower's better} & \multicolumn{1}{c}{higher's better} \\
    \hline
    
    \multirow{4}{*}{SOTA} & Monodepth2 & \multicolumn{1}{c|}{0.303} & \multicolumn{1}{c|}{5.045} & \multicolumn{1}{c}{0.458} \\ \cline{2-5}
    
     & SRNSD \cite{cong2024srnsd} & \multicolumn{1}{c|}{0.169} & \multicolumn{1}{c|}{6.439} & \multicolumn{1}{c}{0.768} \\ \cline{2-5}
    
     & md4All-AD & \multicolumn{1}{c|}{0.123} & \multicolumn{1}{c|}{3.812} & \multicolumn{1}{c}{0.857} \\ \cline{2-5}
    
     & md4All-DD & \multicolumn{1}{c|}{0.123} & \multicolumn{1}{c|}{3.664} & \multicolumn{1}{c}{0.848} \\ \hline
    
    \multirow{4}{*}{\makecell[l]{Plugin\\Enhanced\\Models}} & PhysDepth+DepthDark & \multicolumn{1}{c|}{0.122} & \multicolumn{1}{c|}{3.666} & \multicolumn{1}{c}{0.875} \\ \cline{2-5}
    
     & PhysDepth+ConvNeXt & \multicolumn{1}{c|}{0.122} & \multicolumn{1}{c|}{3.665} & \multicolumn{1}{c}{0.876} \\ \cline{2-5}
    
     & PhysDepth+S2R & \multicolumn{1}{c|}{0.123} & \multicolumn{1}{c|}{3.662} & \multicolumn{1}{c}{0.879} \\ \cline{2-5}
    
     & PhysDepth+ViT-B/16 & \multicolumn{1}{c|}{0.121} & \multicolumn{1}{c|}{3.663} & \multicolumn{1}{c}{0.878} \\ \hline
    
    Ours & \textbf{PhysDepth} & \multicolumn{1}{c|}{\cellcolor{db}\textbf{0.120}} & \multicolumn{1}{c|}{\cellcolor{db}\textbf{3.659}} & \multicolumn{1}{c}{\cellcolor{db}\textbf{0.881}} \\
    
    \bottomrule
    
    \end{tabular}
    }
    \label{tab:60m robot car}
    \end{table}

\begin{table}[ht]
    \centering
    \footnotesize
    \caption{\textbf{Ablation Study:} PhysDepth and Plugin-enhanced Models on all images from the RobotCar-Night dataset \cite{wang2021regularizing} and set depth range up to 80 meters. The best results are in \textbf{bold}.}    \vspace{-2mm}
    \resizebox{\columnwidth}{!}{
    \begin{tabular}{l|l|cc|c}
    \toprule
    \multicolumn{1}{c|}{} & \multicolumn{1}{c|}{Methods} & \multicolumn{1}{c|}{Abs Rel} & \multicolumn{1}{c|}{RMSE} & \multicolumn{1}{c}{$\delta<1.25$} \\
    \multicolumn{1}{c|}{} & \textbf{} & \multicolumn{2}{c|}{lower's better} & \multicolumn{1}{c}{higher's better} \\
    \hline
    
    \multirow{3}{*}{SOTA} & Monodepth2 & \multicolumn{1}{c|}{0.399} & \multicolumn{1}{c|}{6.641} & \multicolumn{1}{c}{0.744} \\ \cline{2-5}
    
     & md4All-AD & \multicolumn{1}{c|}{0.124} & \multicolumn{1}{c|}{3.880} & \multicolumn{1}{c}{0.857} \\ \cline{2-5}
    
     & md4All-DD & \multicolumn{1}{c|}{0.123} & \multicolumn{1}{c|}{3.713} & \multicolumn{1}{c}{0.849} \\ \hline
    
    \multirow{4}{*}{\makecell[l]{Plugin\\Enhanced\\Models}} & PhysDepth+DepthDark & \multicolumn{1}{c|}{\textbf{0.121}} & \multicolumn{1}{c|}{3.659} & \multicolumn{1}{c}{0.877} \\ \cline{2-5}
    
     & PhysDepth+ConvNeXt & \multicolumn{1}{c|}{\textbf{0.121}} & \multicolumn{1}{c|}{3.662} & \multicolumn{1}{c}{0.874} \\ \cline{2-5}
    
     & PhysDepth+S2R & \multicolumn{1}{c|}{\textbf{0.121}} & \multicolumn{1}{c|}{3.659} & \multicolumn{1}{c}{0.876} \\ \cline{2-5}
    
     & PhysDepth+ViT-B/16 & \multicolumn{1}{c|}{\textbf{0.121}} & \multicolumn{1}{c|}{3.658} & \multicolumn{1}{c}{0.877} \\ \hline
    
    Ours & \textbf{PhysDepth} & \multicolumn{1}{c|}{\cellcolor{db}\textbf{0.121}} & \multicolumn{1}{c|}{\cellcolor{db}\textbf{3.657}} & \multicolumn{1}{c}{\cellcolor{db}\textbf{0.879}} \\
    
    \bottomrule
    
    \end{tabular}
    }
    \label{tab:80m robot car}
    \end{table}

Table \ref{tab:40m nuScenes}, \ref{tab:50m nuScenes}, and \ref{tab:60m nuScenes} are the results of nuScenes-Night dataset \cite{caesar2020nuscenes} with different depth scales. Specifically, we alter the $\beta$ value  from 80 meters to 40, 50, and 60 meters, respectively.

The results show that PhysDepth outperforms SOTA methods in all distances, demonstrating PhysDepth's robustness under challenging environments. Additionally, the plugin (PPM + Loss Functions) effectively enhanced existing MDE models.

Table \ref{tab:40m robot car}, \ref{tab:60m robot car}, and \ref{tab:80m robot car} show the results of RobotCar-Night dataset \cite{RobotCarDatasetIJRR} with depth scales of 40, 60, and 80 meters, respectively. The results affirm that \textit{PhysDepth can accurately estimate the depth under challenging environments despite the varying depth scales}. Notably, in Table \ref{tab:40m robot car}, we significantly outperform Monodepth2 in RMSE. 

\subsection{Additional Ablation Study}

\begin{table}[ht]
    \centering
    \footnotesize
    \caption{\textbf{Ablation Study of $f(R)$ and $R$:} Using red channel images from the RobotCar-Night dataset \cite{wang2021regularizing} and the depth range is up to 80m. The best results are in \textbf{bold}.}    \vspace{-2mm}
    \resizebox{\columnwidth}{!}{
    \begin{tabular}{l|cc|c}
    \toprule
    \multicolumn{1}{c|}{Methods} & \multicolumn{1}{c|}{Abs Rel} & \multicolumn{1}{c|}{RMSE} & \multicolumn{1}{c}{$\delta<1.25$} \\
    \multicolumn{1}{c|}{} & \multicolumn{2}{c|}{lower's better} & \multicolumn{1}{c}{higher's better} \\
    \hline
    
    PhysDepth+$R$ & \multicolumn{1}{c|}{0.147} & \multicolumn{1}{c|}{3.981} & \multicolumn{1}{c}{0.635} \\ \hline
    
    PhysDepth+$f(R)$ & \multicolumn{1}{c|}{\cellcolor{db}\textbf{0.121}} & \multicolumn{1}{c|}{\cellcolor{db}\textbf{3.657}} & \multicolumn{1}{c}{\cellcolor{db}\textbf{0.879}} \\
    
    \bottomrule
    
    \end{tabular}
    }
    \label{tab:r and fr}
    \end{table}

In Table \ref{tab:r and fr}
, we use Equation \ref{eq:rca loss} and Equation \ref{eq:rca loss r and c} to estimate $d_R$ respectively.

PhysDepth + $R$ shows the result from Equation \ref{eq:rca loss}. 
And PhysDepth + $f(R)$ shows the result from Equation \ref{eq:rca loss r and c}.
These results further demonstrating using $f(R)$ is better that $R$ in our model becasue $f(R)$ contains low and high level feature from image. 

\begin{figure*}[ht]
	\centering
	\includegraphics[width=0.8\linewidth]{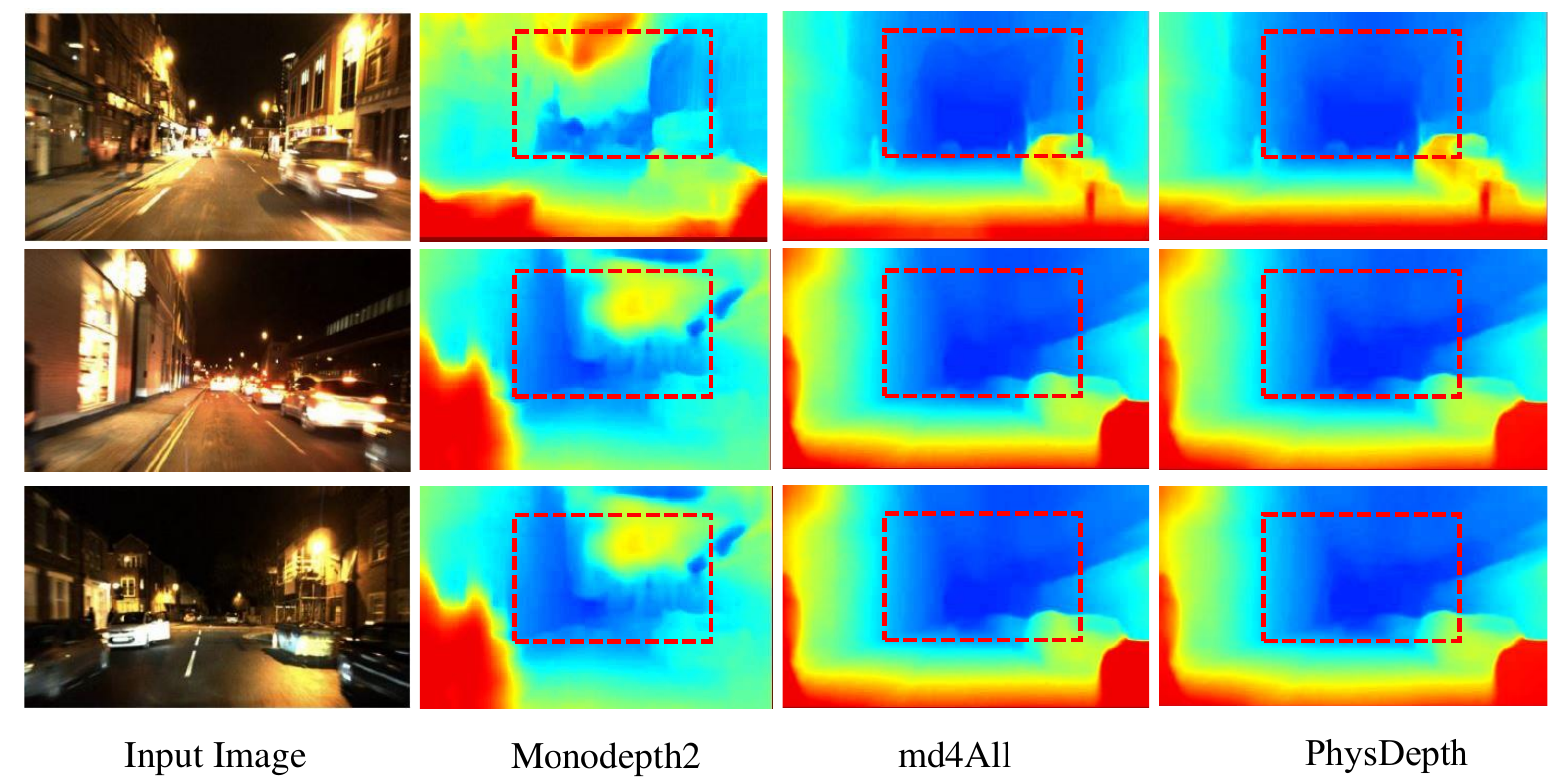}
	\caption{\textbf{Qualitative Results - RobotCar-Night:} Depth estimated on three different test images (column: Input Image) using our approach (column: PhysDepth) in comparison with two other methods (remaining columns).}
	\label{pic:RobotCar-Night visual sup}
\end{figure*}

\begin{equation}
\label{eq:linear}
    I = I_0 \frac{1}{a \times d_R + b}
\end{equation}

\begin{equation}
\label{eq:quadratic}
    I = I_0 \frac{1}{a \times d_R^{2} + b \times d_R + c }
\end{equation}

\begin{table}[ht]
    \centering
    \footnotesize
    \caption{\textbf{Ablation Study of $\mu$:} Using different $\mu$ on RobotCar-Night dataset \cite{wang2021regularizing} and the depth range is up to 80m.  The best results are in \textbf{bold}.}    \vspace{-2mm}
    \resizebox{\columnwidth}{!}{
    \begin{tabular}{l|cc|c}
    \toprule
    \multicolumn{1}{c|}{Methods} & \multicolumn{1}{c|}{Abs Rel} & \multicolumn{1}{c|}{RMSE} & \multicolumn{1}{c}{$\delta<1.25$} \\
    \multicolumn{1}{c|}{} & \multicolumn{2}{c|}{lower's better} & \multicolumn{1}{c}{higher's better} \\
    \hline
    
    $\mu$=0.1 & \multicolumn{1}{c|}{0.139} & \multicolumn{1}{c|}{4.875} & \multicolumn{1}{c}{0.315} \\ \hline
    
    $\mu$=0.05 & \multicolumn{1}{c|}{0.125} & \multicolumn{1}{c|}{4.362} & \multicolumn{1}{c}{0.741} \\ \hline
    
    $\mu$=0.025 & \multicolumn{1}{c|}{0.120} & \multicolumn{1}{c|}{4.257} & \multicolumn{1}{c}{0.853} \\ \hline
    
    \textbf{Learnable $\mu$} & \multicolumn{1}{c|}{\cellcolor{db}\textbf{0.121}} & \multicolumn{1}{c|}{\cellcolor{db}\textbf{3.657}} & \multicolumn{1}{c}{\cellcolor{db}\textbf{0.879}} \\
    
    \bottomrule
    
    \end{tabular}
    }
    \label{tab:Ablation Study of mu}
    \end{table}

Table \ref{tab:Ablation Study of mu} presents the ablation study with different strategies to choose $\mu$ in Equation \ref{eq:Beer-Lambert law}; For the first strategy, we set $\mu$ to a fixed value. To ensure a fair comparison, we respectively set $\mu$ to $0.05$, $0.025$, and $0.0125$, which are selected based on 80 meters' depth range.
For the second strategy, we set $\mu$ as a learnable parameter, which is the implementation of PhysDepth. 

\begin{table}[ht]
    \centering
    \footnotesize   
    \caption{\textbf{Ablation Study of $\lambda$:} Using different $\lambda$ on RobotCar-Night dataset \cite{wang2021regularizing} and the depth range is up to 80m.  The best results are in \textbf{bold}.}    \vspace{-2mm}
    \resizebox{\columnwidth}{!}{
    \begin{tabular}{l|cc|c}
    \toprule
    \multicolumn{1}{c|}{Methods} & \multicolumn{1}{c|}{Abs Rel} & \multicolumn{1}{c|}{RMSE} & \multicolumn{1}{c}{$\delta<1.25$} \\
    \multicolumn{1}{c|}{} & \multicolumn{2}{c|}{lower's better} & \multicolumn{1}{c}{higher's better} \\
    \hline
    
    $\lambda$=0.01 & \multicolumn{1}{c|}{0.358} & \multicolumn{1}{c|}{9.182} & \multicolumn{1}{c}{0.980} \\ \hline
    
    $\lambda$=0.1 & \multicolumn{1}{c|}{0.227} & \multicolumn{1}{c|}{4.775} & \multicolumn{1}{c}{0.677} \\ \hline
    
    $\lambda$=0.003 & \multicolumn{1}{c|}{0.195} & \multicolumn{1}{c|}{3.951} & \multicolumn{1}{c}{0.451} \\ \hline
    
    \textbf{Learnable $\lambda$} & \multicolumn{1}{c|}{\cellcolor{db}\textbf{0.121}} & \multicolumn{1}{c|}{\cellcolor{db}\textbf{3.657}} & \multicolumn{1}{c}{\cellcolor{db}\textbf{0.879}} \\
    
    \bottomrule
    
    \end{tabular}
    }
    \label{tab:Ablation Study of lambda}
\end{table}

According to Table \ref{tab:Ablation Study of mu} and \ref{tab:Ablation Study of lambda},  setting $\mu$ and $\lambda$ as learnable parameters gives the best accuracy. 
This is because each pixel of the same input image may represent a unique light attenuation condition; dynamically assigning distinct $\mu$ and $\lambda$ values to each pixel 
enables each pixel to automatically adjust to its own light attenuation conditions. 

To determine the optimal value for the parameter $g$ in Equation \ref{eq:sg}, we conducted empirical tests using various values. Our results, presented in Table \ref{tab:Ablation Study of g}, indicate that setting $g$ to 1.3938 yields the highest accuracy.

We also test PhysDepth with multiple views (3, and 5 views) as input. However, the results exhibit negligible improvements, aligning with the conclusions of Ummenhofer et al. \cite{ummenhofer2017demon} and Zhou et al. \cite{zhou2017unsupervised}.

\begin{table}[ht]
    \centering
    \footnotesize
    \caption{\textbf{Ablation Study of gg:} Using different gg on RobotCar-Night dataset \cite{wang2021regularizing} and the depth range is up to 80m.  The best results are in \textbf{bold}.}    \vspace{-2mm}
    \resizebox{\columnwidth}{!}{
    \begin{tabular}{l|cc|c}
    \toprule
    \multicolumn{1}{c|}{Methods} & \multicolumn{1}{c|}{Abs Rel} & \multicolumn{1}{c|}{RMSE} & \multicolumn{1}{c}{$\delta<1.25$} \\
    \multicolumn{1}{c|}{} & \multicolumn{2}{c|}{lower's better} & \multicolumn{1}{c}{higher's better} \\
    \hline
    
    $g$=0.5 & \multicolumn{1}{c|}{0.921} & \multicolumn{1}{c|}{6.175} & \multicolumn{1}{c}{0.710} \\ \hline
    
    $g$=0.1 & \multicolumn{1}{c|}{0.550} & \multicolumn{1}{c|}{5.587} & \multicolumn{1}{c}{0.515} \\ \hline
    
    $g$=0.01 & \multicolumn{1}{c|}{0.515} & \multicolumn{1}{c|}{5.251} & \multicolumn{1}{c}{0.355} \\ \hline
    
    $g$=1.5 & \multicolumn{1}{c|}{0.873} & \multicolumn{1}{c|}{5.891} & \multicolumn{1}{c}{0.689} \\ \hline
    
    $g$=1.3938 & \multicolumn{1}{c|}{\cellcolor{db}\textbf{0.121}} & \multicolumn{1}{c|}{\cellcolor{db}\textbf{3.657}} & \multicolumn{1}{c}{\cellcolor{db}\textbf{0.879}} \\
    
    \bottomrule
    
    \end{tabular}
    }
    \label{tab:Ablation Study of g}
\end{table}

\begin{figure*}[ht]
    \centering
    \includegraphics[width=0.8\linewidth]{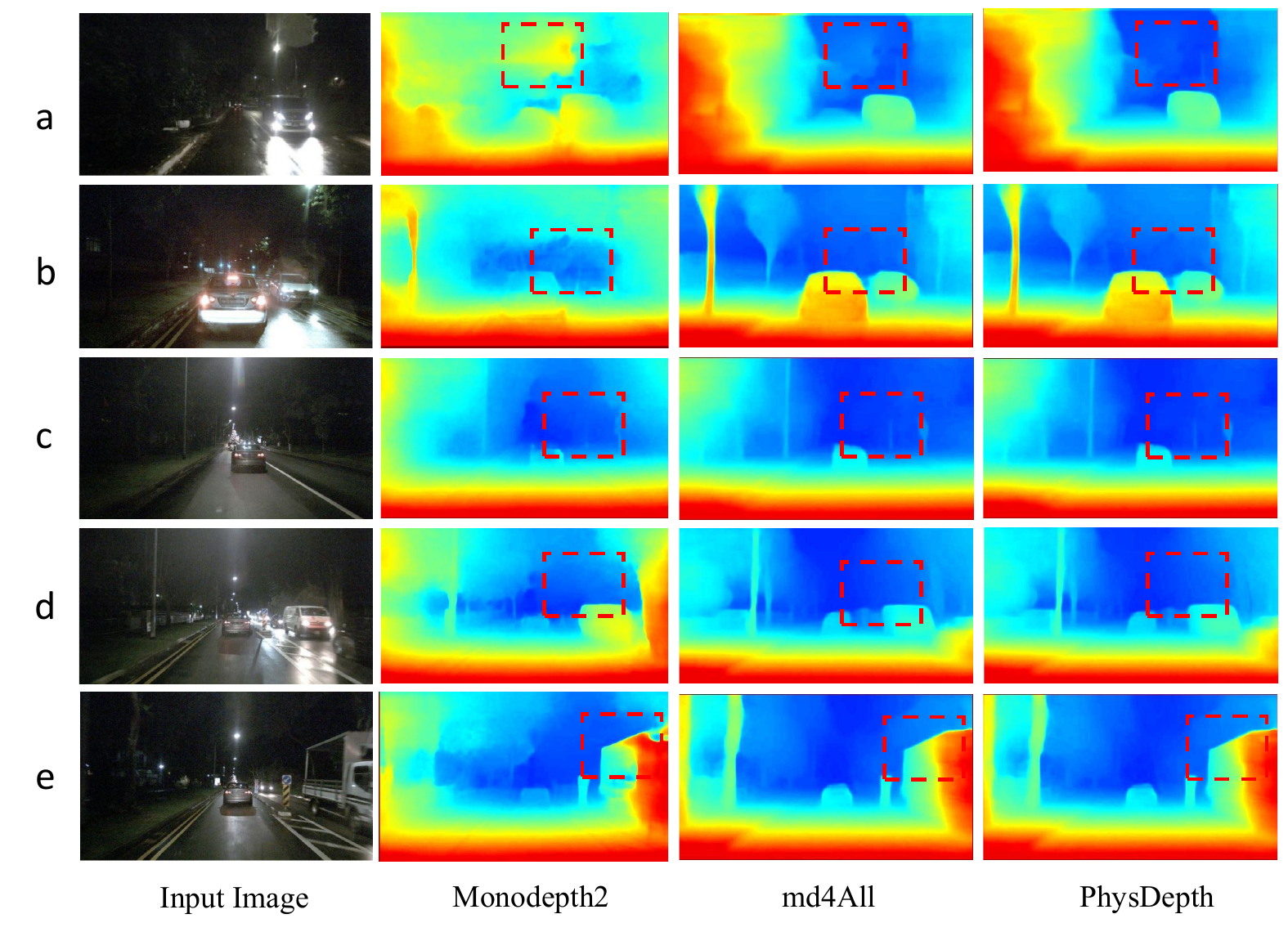}
    \caption{\textbf{Qualitative Results - nuScence-Night:} Depth estimated on five different test images (column: Input) using our approach (column: PhysDepth) in comparison with two SOTA methods (remaining columns).}
    \label{pic:nuScence visual sup}
\end{figure*}

\begin{table}[ht]
    \centering
    \scriptsize
    \caption{\textbf{Quantitative Results - KITTI \cite{geiger2012we}:} Comparison of our model on the KITTI test split by Eigen \cite{eigen2014depth}. M means monocular image, and S means stereo image.
    The best results are in \textbf{bold} for each metric. All results presented here are without post-processing.}
    \vspace{-3mm}
    \resizebox{\columnwidth}{!}{
    \begin{tabular}{l|c|c|cc|c}
    \toprule
    \multicolumn{1}{c|}{\multirow{2}{*}{Methods}} & \multicolumn{1}{l|}{\multirow{2}{*}{Data}} & \multirow{2}{*}{$H \times W$} & \multicolumn{1}{c|}{Abs Rel} & \multicolumn{1}{c|}{RMSE} & \multicolumn{1}{c}{$\delta<1.25$} \\
    \cline{4-6} 
    \multicolumn{1}{c|}{}                         & \multicolumn{1}{l|}{} &  & \multicolumn{2}{c|}{lower's better} & \multicolumn{1}{c}{higher's better} \\
    \hline
    DynaDepth \cite{feng2022disentangling} & S & $192 \times 640$ & \multicolumn{1}{c|}{0.109} & \multicolumn{1}{c|}{4.705} & \multicolumn{1}{c}{0.869} \\ \cline{1-1} \cline{4-6} 
    {MonoDepth2 \cite{godard2019digging}} & S & $320 \times 1024$ & \multicolumn{1}{c|}{0.107} & \multicolumn{1}{c|}{4.764} & \multicolumn{1}{c}{0.874} \\ \cline{1-1} \cline{4-6} 
    EPCDepth \cite{peng2021excavating} & S & $320 \times 1024$ & \multicolumn{1}{c|}{0.091} & \multicolumn{1}{c|}{4.207} & \multicolumn{1}{c}{0.901} \\ \cline{1-1} \cline{4-6} 
    DepthFormer \cite{li2022depthformer} & S & $320 \times 1024$ & \multicolumn{1}{c|}{0.090} & \multicolumn{1}{c|}{4.149} & \multicolumn{1}{c}{0.905} \\ \cline{1-1} \cline{4-6} 
    PlaneDepth \cite{wang2023planedepth} & S & $384 \times 1280$ & \multicolumn{1}{c|}{0.083} & \multicolumn{1}{c|}{3.919} & \multicolumn{1}{c}{0.913} \\ \cline{1-1} \cline{4-6} 
    \textbf{PhysDepth} & S & \multicolumn{1}{l|}{$320 \times 1024$} & \multicolumn{1}{c|}{\cellcolor{db}\textbf{0.079}} & \multicolumn{1}{c|}{\cellcolor{db}\textbf{3.915}} & \multicolumn{1}{c}{\cellcolor{db}\textbf{0.917}} \\ \hline
    \hline
    SfMLearner \cite{zhou2017unsupervised} & M & $128 \times 416$ & \multicolumn{1}{c|}{0.198} & \multicolumn{1}{c|}{6.565} & \multicolumn{1}{c}{0.718} \\ \cline{1-1} \cline{4-6} 
    DeFeat-Net \cite{spencer2020defeat} & M & $352 \times 480$ & \multicolumn{1}{c|}{0.126} & \multicolumn{1}{c|}{5.035} & \multicolumn{1}{c}{0.862} \\ \cline{1-1} \cline{4-6} 
    CADepth-Net \cite{yan2021channel} & M & $192 \times 640$ & \multicolumn{1}{c|}{0.105} & \multicolumn{1}{c|}{4.535} & \multicolumn{1}{c}{0.892} \\ \cline{1-1} \cline{4-6} 
    FM \cite{shu2020feature} & M & $192 \times 640$ & \multicolumn{1}{c|}{0.104} & \multicolumn{1}{c|}{4.481} & \multicolumn{1}{c}{0.893} \\ \cline{1-1} \cline{4-6} 
    GCP-MonoViT \cite{moon2024ground} & M & $192 \times 640$ & \multicolumn{1}{c|}{0.096} & \multicolumn{1}{c|}{4.327} & \multicolumn{1}{c}{0.904} \\ \cline{1-1} \cline{4-6} 
    \textbf{PhysDepth} & M & $192 \times 640$ & \multicolumn{1}{c|}{\cellcolor{db}\textbf{0.093}} & \multicolumn{1}{c|}{\cellcolor{db}\textbf{4.013}} & \multicolumn{1}{c}{\cellcolor{db}\textbf{0.906}} \\
    \hline
    \hline
    {MonoDepth2 \cite{godard2019digging}} & M & $320 \times 1024$ & \multicolumn{1}{c|}{0.115} & \multicolumn{1}{c|}{4.701} & \multicolumn{1}{c}{0.879} \\ \cline{1-1} \cline{4-6} 
    DynamicDepth\cite{feng2022disentangling} & M & $192 \times 640$ & \multicolumn{1}{c|}{0.103} & \multicolumn{1}{c|}{5.867} & \multicolumn{1}{c}{0.897} \\ \cline{1-1} \cline{4-6} 
    CADepth-Net \cite{yan2021channel} & M & $320 \times 1024$ & \multicolumn{1}{c|}{0.102} & \multicolumn{1}{c|}{4.407} & \multicolumn{1}{c}{0.898} \\ \cline{1-1} \cline{4-6} 
    DIFFNet \cite{park2021diffnet} & M & $320 \times 1024$ & \multicolumn{1}{c|}{0.097} & \multicolumn{1}{c|}{4.345} & \multicolumn{1}{c}{0.907} \\ \cline{1-1} \cline{4-6} 
    \textbf{PhysDepth} & M & $320 \times 1024$ & \multicolumn{1}{c|}{\cellcolor{db}\textbf{0.095}} & \multicolumn{1}{c|}{\cellcolor{db}\textbf{3.995}} & \multicolumn{1}{c}{\cellcolor{db}\textbf{0.912}} \\
    \bottomrule
    \end{tabular}
    }
    \label{tab:KITTI quan}
    \end{table}
    \section{Additional Qualitative Results}
\label{sec:Additional Qualitative Results}

\begin{figure*}[ht]
	\centering
	\includegraphics[width=1\linewidth]{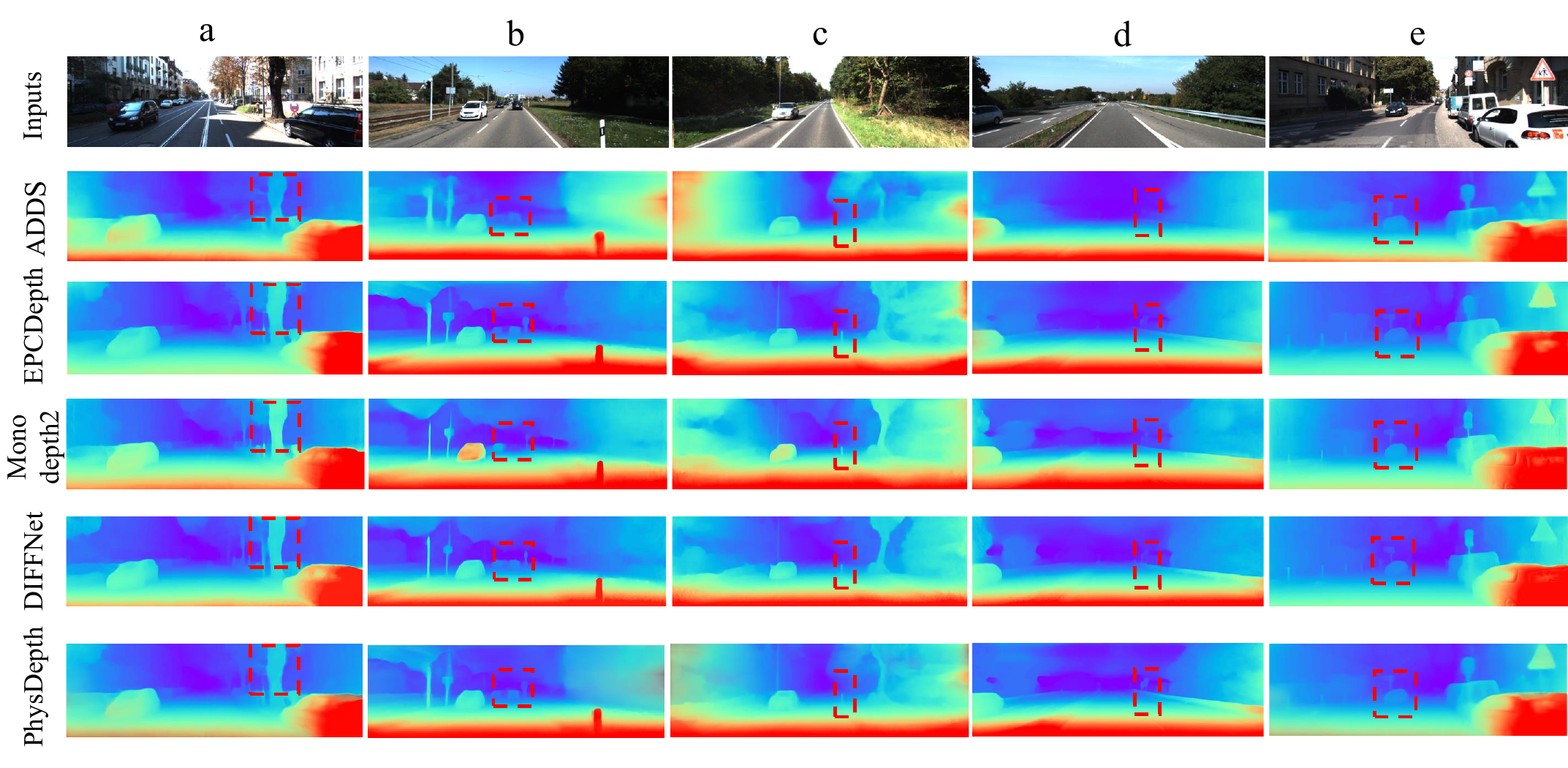}
	\vspace{-6mm}
	\caption{\textbf{Qualitative Results - KITTI:} Depth estimated on five different test images (row: Inputs) using our model (row: PhysDepth) and four SOTA (remaining rows) for comparison.}
	\vspace{-3mm}
	\label{fig:KITTI visual}
\end{figure*}

Figure \ref{pic:RobotCar-Night visual sup} shows the results on the RobotCar-Night dataset, demonstrating that our model provides more accurate MDE results compared to others.  
For instance, The red boxes in Figure \ref{pic:RobotCar-Night visual sup} - a,b,c

demonstrate our model's ability to accurately handle varying lighting conditions. 

In Figures \ref{pic:nuScence visual sup}, we use nuScence-Night dataset to compare PhysDepth with SOTA methods: Monodepth2 \cite{godard2019digging} and md4All \cite{gasperini2023robust}.
From the results, PhysDepth shows sharper details for the roads and cars, showcasing PhysDepth's ability to accurately handle varying light conditions under challenging scenarios. 

Notably, the input images are particularly challenging as they were taken in extremely dark environments. However, PhysDepth can still provide robust results under those extreme conditions.

\section{Evaluation on Daytime Dataset}
\label{sec:daytime}
We use KITTI \cite{geiger2012we} dataset to verify the performance of PhysDepth under daytime environments.
Table \ref{tab:KITTI quan} highlights that our model outperforms all SOTA models across all metrics, emphasizing its overall generality and accuracy under daytime environments.

Figure \ref{fig:KITTI visual} qualitatively demonstrates that PhysDepth, while designed for MDE under challenging conditions, outperforms SOTA models under daytime scenarios. 
For example, in Figure \ref{fig:KITTI visual} - b, our model accurately detects the car, providing clear and sharp boundaries. In contrast, MonoDepth2 fails to detect the car entirely. While ADDS, EPCDepth, and DIFFNet can locate the car, the boundaries they produce are noticeably blurred, resulting in the two cars being jammed together on the estimated depth maps.
\textit{The physical priors employed by PhysDepth are environment-invariant, ensuring robust performance in both daytime scenarios and in challenging conditions.}.

\begin{figure*}[ht]
    \centering
    \includegraphics[width=0.8\linewidth]{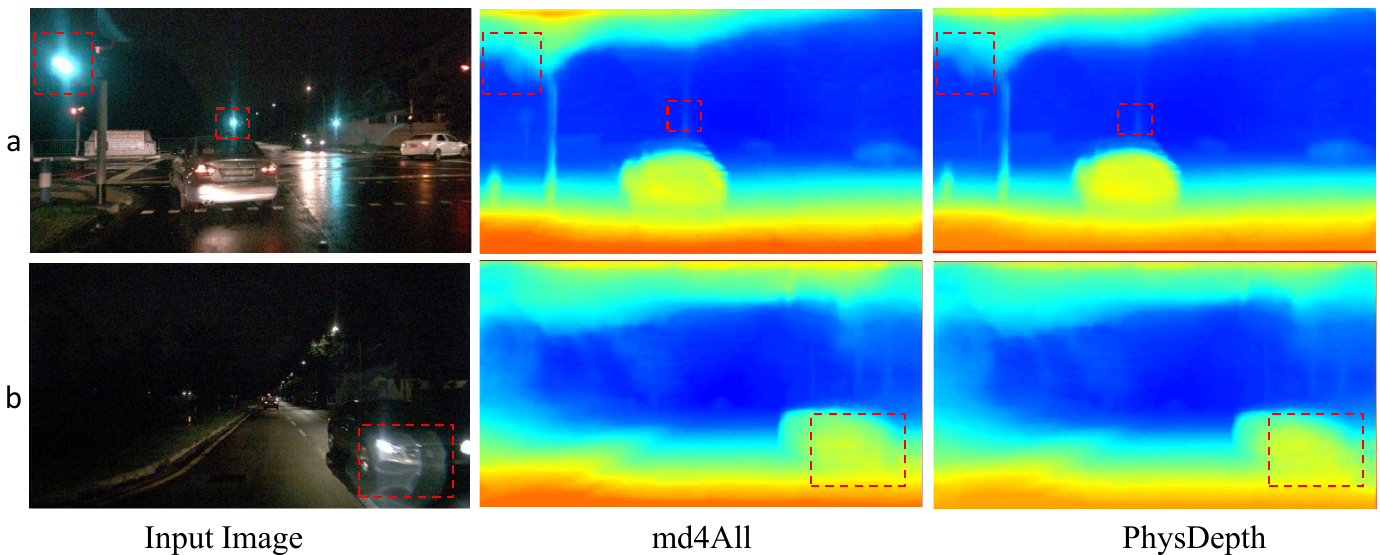}
    \caption{\textbf{Failure Cases - nuScence-Night:} Comparison between PhysDepth and md4All \cite{gasperini2023robust}. Red boxes show the halos because of road lamps and car headlights in the input image, which may cause failure for both PhysDepth and md4All.}
    \label{pic:failure case nuScence}
\end{figure*}

\section{Failure Cases}
\label{sec:Failure Cases}
PhysDepth performs well under challenging environments, but there's room for improvement with specific corner cases.
For example, Figure \ref{pic:failure case nuScence} shows two instances of the failure cases. 

In the input images, halos originating from road lamps and other car headlights are the main reasons for these failure cases.
These halos give rise to shadows and reflections, which could affect the light attenuation item (Equation \ref{eq:Beer-Lambert law}) in RCA loss, leading to compromised depth estimation results.

The halo issue is commonly known as one of the most challenging conditions in nighttime depth estimation \cite{wang2021regularizing}. 
From Figure \ref{pic:failure case nuScence}, we can see that SOTA method md4All \cite{gasperini2023robust} produces similar depth results as PhysDepth, demonstrating that SOTA methods may not be able to handle the failure case well.

\end{document}